\documentclass{article} % For LaTeX2e
\usepackage{hunyuan,times}
\usepackage{lipsum} % 用于生成示例文字

\usepackage{amsmath,amsfonts,bm}

\def\eqref#1{equation~\ref{#1}}
\def\1{\bm{1}}

\DeclareMathAlphabet{\mathsfit}{\encodingdefault}{\sfdefault}{m}{sl}
\SetMathAlphabet{\mathsfit}{bold}{\encodingdefault}{\sfdefault}{bx}{n}

\usepackage{multirow}      % 用于创建跨多行的单元格
\usepackage{amssymb}       % 提供了丰富的数学符号，包括 \checkmark (对勾)
\usepackage{graphicx}     % 用于处理图片和缩放内容 (\resizebox)

\usepackage{colortbl}
\usepackage{xcolor}

\usepackage{hyperref}
\usepackage{url}
\usepackage{multicol}
\usepackage{aliascnt}
\usepackage{adjustbox}
\usepackage{tcolorbox}
\usepackage{siunitx} % for S column type in tables
\usepackage{multirow} 
\usepackage{enumitem}
\usepackage{amssymb}  % 用于 \checkmark
\usepackage{amsmath}  % 用于 \texttimes (或者使用 pifont 和 \ding{55})
\usepackage{longtable} % 如果表格太长需要跨页
\usepackage{graphicx}
\usepackage{subcaption}
\usepackage{calc}

\usepackage{booktabs}
\usepackage{titletoc}
\usepackage{wrapfig}
\usepackage[dvipsnames, svgnames, table]{xcolor}

\colorlet{offpolicycolor}{orange!5}
\colorlet{onpolicycolor}{blue!10}

\newcommand{\I}{\mathbb{I}}
\usepackage{bbding}

\usepackage{CJKutf8} % 支持中文
\definecolor{uclablue}{rgb}{0.15, 0.45, 0.68}
\hypersetup{
    breaklinks,
    citecolor=uclablue,
    colorlinks=true,
}

\newtcolorbox{AIbox}[2][]{aibox,title=#2,#1}
\tcbset{
  aibox/.style={
    top=10pt,
    colframe=black,
    colbacktitle=black,
    center
  }
}

\title{Low-probability Tokens Sustain Exploration in Reinforcement Learning with Verifiable Reward}

\author{
Guanhua Huang$^{1,}$\thanks{\ Equal contribution.}~~, Tingqiang Xu$^{1,2,*}$\thanks{\ Work completed during an internship at Tencent.}~~, Mingze Wang$^{1,3,\dagger}$, Qi Yi$^1$, Xue Gong$^1$, Siheng Li$^{1,4,\dagger}$,\\ Ruibin Xiong$^1$, 
Kejiao Li$^1$, Yuhao Jiang$^1$, Bo Zhou${^1}$\thanks{\ Correspondence to Bo Zhou: chaysezhou@tencent.com.} \\
\vspace{2mm}
\textbf{$^1$LLM Department, Tencent} \quad
\textbf{$^2$Tsinghua University} \quad\\
\textbf{$^3$Peking University} \quad
\textbf{$^4$The Chinese University of Hong Kong}
}

\begin{document}
\maketitle
\let\oldthefootnote\thefootnote

\let\thefootnote\oldthefootnote

\begin{abstract}
Reinforcement Learning with Verifiable Rewards (RLVR) has propelled Large Language Models in complex reasoning, yet its scalability is often hindered by a training bottleneck where performance plateaus as policy entropy collapses, signaling a loss of exploration. Previous methods typically address this by maintaining high policy entropy, yet the precise mechanisms that govern meaningful exploration have remained underexplored. Our analysis suggests that an unselective focus on entropy risks amplifying irrelevant tokens and destabilizing training. This paper investigates the exploration dynamics within RLVR and identifies a key issue: the gradual elimination of valuable low-probability exploratory tokens, which we term \textbf{\textit{reasoning sparks}}. We find that while abundant in pre-trained models, these sparks are systematically extinguished during RLVR due to over-penalization, leading to a degeneracy in exploration. To address this, we introduce Low-probability Regularization (Lp-Reg). Its core mechanism regularizes the policy towards a heuristic proxy distribution. This proxy is constructed by filtering out presumed noise tokens and re-normalizing the distribution over the remaining candidates. The result is a less-noisy proxy where the probability of \textit{reasoning sparks} is amplified, which then serves as a soft regularization target to shield these valuable tokens from elimination via KL divergence. Experiments show that Lp-Reg enables stable on-policy RL, sustaining continuous scaling across $3,000$ training steps and $81,204$ GPU-hours, where baseline entropy-control methods collapse. This sustained exploration leads to state-of-the-art performance, achieving a $60.17\%$ average accuracy on five math benchmarks, an improvement of $2.66\%$ over prior methods. Code is available at \url{https://github.com/CarlanLark/Lp-Reg}.
\end{abstract}

\section{Introduction}

The advent of large reasoning models has reshaped the trajectory of artificial intelligence, with paradigmatic examples including OpenAI O1~\citep{openai2024openaio1card} and DeepSeek-R1~\citep{deepseekai2025deepseekr1incentivizingreasoningcapability}. A central technique underpinning these systems is reinforcement learning with verifiable reward (RLVR), which assigns reward to verifiable solutions through rule-based verification. These models generate extended chain-of-thought (CoT) reasoning~\citep{wei2023chainofthoughtpromptingelicitsreasoning} to solve challenging problems in domains like mathematical olympiads~\citep{he2024olympiadbench}. However, a notable bottleneck emerges during RL training that limits its scalability, frequently culminating in a performance plateau and subsequent collapse. This failure is consistently accompanied by a rapid decay in policy entropy, indicating a severe loss of exploration capacity~\citep{yu2025dapoopensourcellmreinforcement, cui2025entropymechanismreinforcementlearning,wang20258020rulehighentropyminority}.

Previous approaches have recognized this declining exploration, attempting to address it through various entropy control mechanisms. Methods such as adaptive entropy regularization~\citep{he2025skyworkopenreasoner1}, high entropy change blocking~\citep{cui2025entropymechanismreinforcementlearning}, or selective token updates~\citep{wang20258020rulehighentropyminority} aim to maintain higher entropy as a proxy for exploration. However, relying on overall entropy can be an indirect and imprecise tool. An indiscriminate focus on maximizing randomness risks amplifying noise and destabilizing training~\citep{2025failuremodesmaximumentropy}, suggesting a deeper issue beyond simply the quantity of randomness.

\begin{figure}[!t]
    \centering
    % \includegraphics[width=1.0\linewidth]{figures/intro2.pdf}
    
    % \begin{minipage}[c]{0.3\linewidth}
    %     \centering
    %     \subfloat[\small Methods comparison.]{\label{fig:method_comparison_table}
    %     \scalebox{0.50}{ 
    %     \begin{tabular}{lcc}
    %     \toprule[1pt]
    %     \multicolumn{1}{l}{\textbf{Methods}} &
    %     \multicolumn{1}{c}{\textbf{\begin{tabular}[c]{@{}c@{}}Low-probability\\ Explotary \\ Tokens \\ Protection\end{tabular}}} &
    %     \multicolumn{1}{c}{\textbf{\begin{tabular}[c]{@{}c@{}}Low-probability\\ Irrelevant \\ Tokens \\ Penalization\end{tabular}}} \\
    %     \midrule[0.5pt]
    %     GRPO           & \myCrossMark & \myCheckMark  \\
    %     Entropy Loss & \myCheckMark & \myCrossMark \\
    %     Lp-Reg (Ours)  & \myCheckMark & \myCheckMark  \\
    %     \bottomrule[1pt]
    %     \end{tabular}
    %     }}
    % \end{minipage}
    % \hspace{0.02\linewidth}
    \begin{minipage}[c]{0.9\linewidth}
    
        \subfloat[\small \textit{Reasoning Sparks}: low-probability exploratory tokens that initiates a new reasoning path.]{\label{fig:a}
        \vspace{-0.02\linewidth}
        \includegraphics[width=\linewidth]{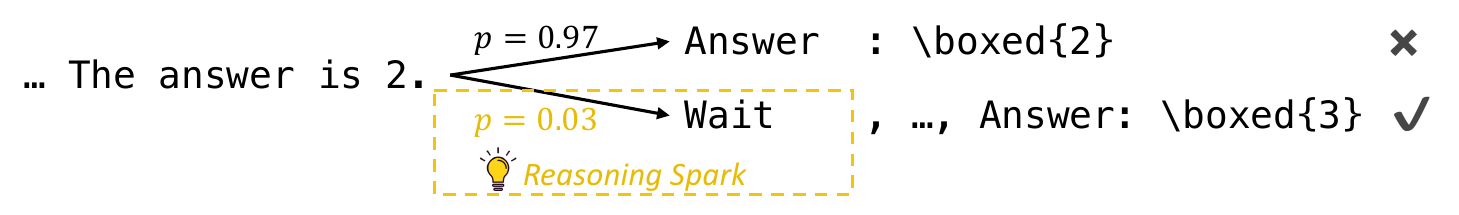}}
    \end{minipage}
    
    \vspace{0.02\linewidth}
    
    \begin{minipage}[c]{0.3\linewidth}
        \centering
        \subfloat[\small Training Dynamics]{\label{fig:b}
        \includegraphics[width=1.0\linewidth]{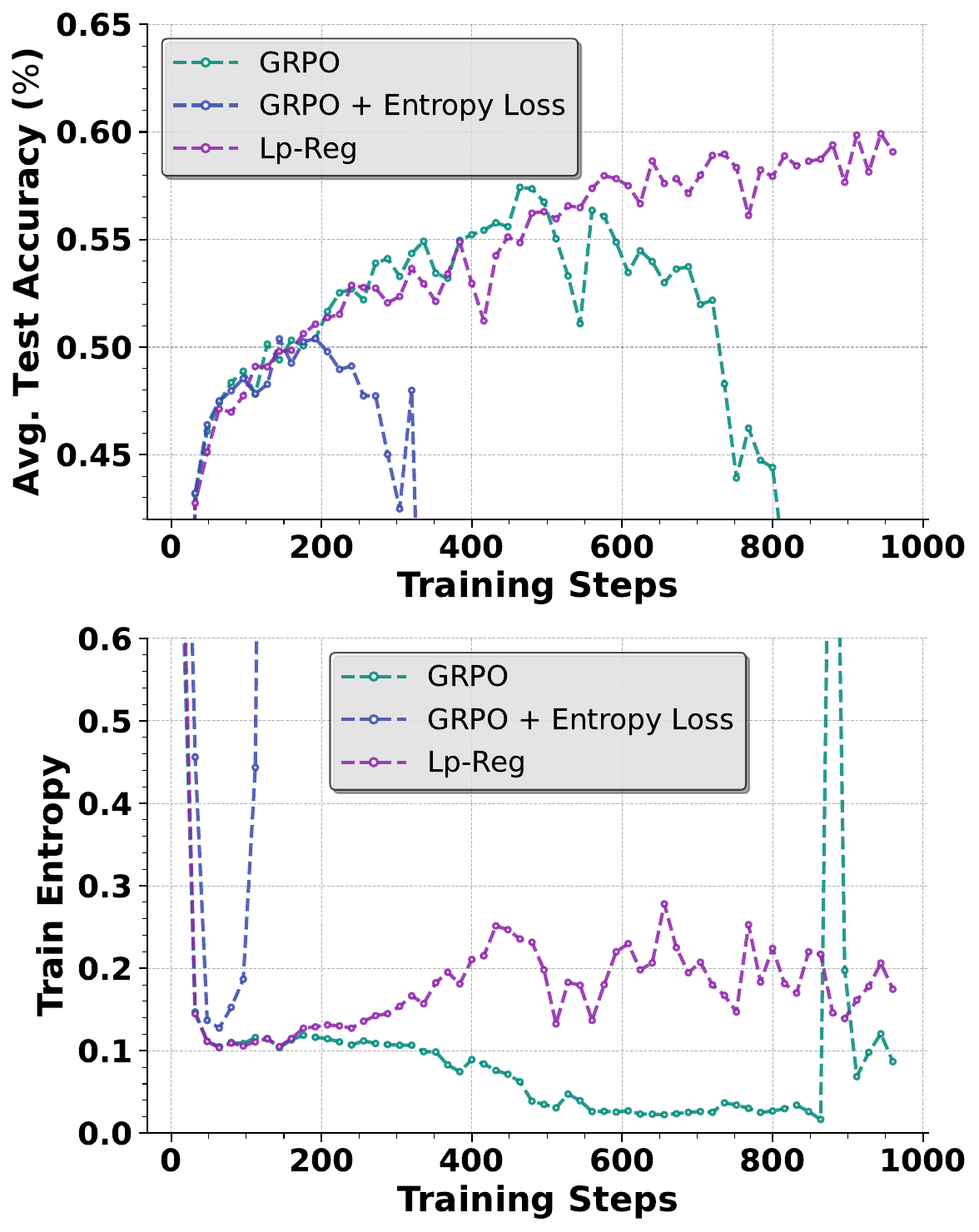}}
    
    \end{minipage}
    \hspace{0.02\linewidth}
    \begin{minipage}[c]{0.65\linewidth}
        \vspace{-0.02\linewidth}
        \subfloat[\small Aggregated distribution of observed sampling probabilities for a class of meaningful exploratory tokens (e.g., ``wait'', ``however''). ``n'' represents sampling number.]{\label{fig:c}
        \includegraphics[width=1.0\linewidth]{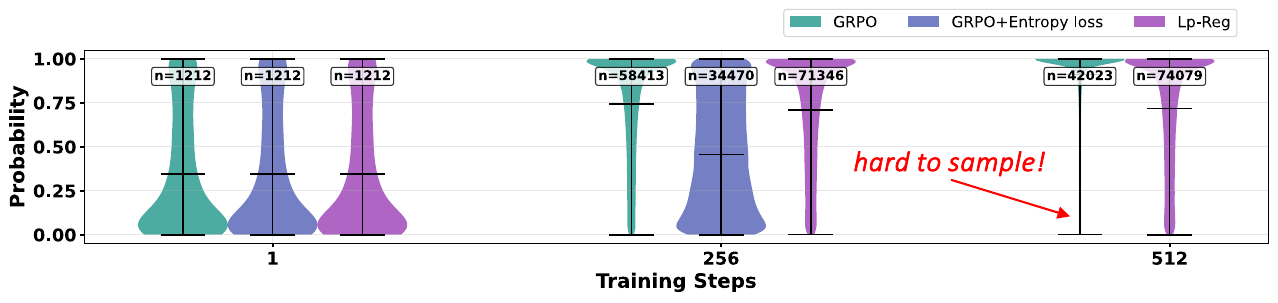}}\\ 
        \subfloat[\small Aggregated distribution of observed sampling probabilities for a class of irrelevant tokens (e.g., ``cost'', ``fine''). ``n'' represents sampling number.]{\label{fig:d}
        \includegraphics[width=1.0\linewidth]{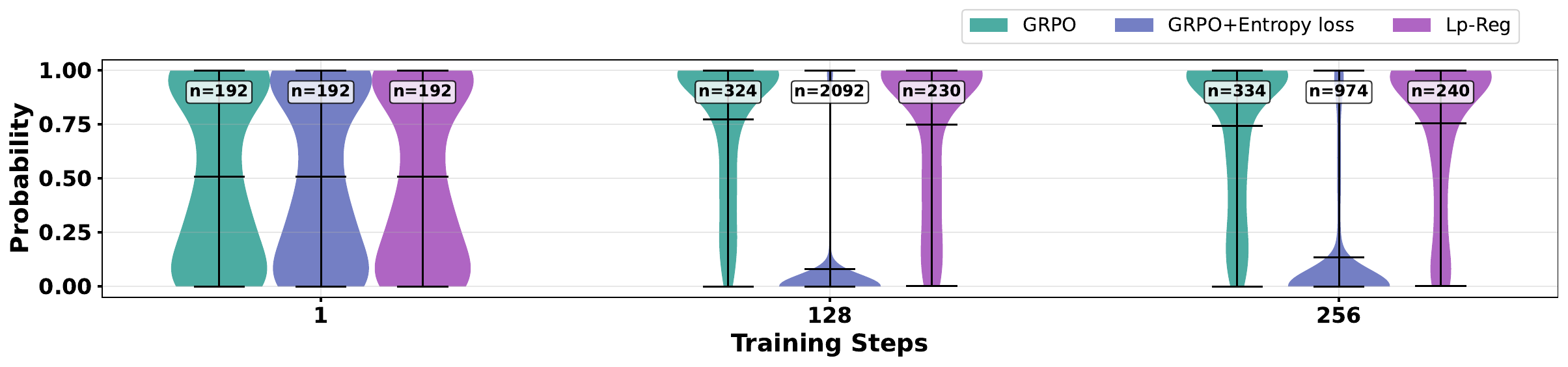}}
    \end{minipage}

    \caption{Selectively preserving low-probability tokens is key to overcoming performance plateaus in reasoning RL. \textbf{(a)} An illustration of a \textit{reasoning spark}. \textbf{(b)} Standard GRPO training reaches a performance plateau and collapses, accompanied by decaying entropy. An indiscriminate entropy bonus (GRPO + Entropy Loss) leads to an even faster collapse. \textbf{(c)} We reveal the cause: GRPO systematically suppresses the low-probability sampling of important exploratory tokens (like ``wait''), and forces these tokens' sampling distributions to collapse towards high probabilities. Entropy Loss fails to fix this. In contrast, our method, Lp-Reg, successfully preserves a healthy, wide distribution, sustaining exploration. \textbf{(d)} The failure of entropy bonuses is explained by amplifying the low-probability sampling of irrelevant tokens, creating noise, and thereby degrading exploration quality. The aggregated statistics in (c) and (d) demonstrate a systemic effect beyond single-token instances. Detailed plots for individual tokens are available in Appendix~\ref{sec:Details of Sampling Probability Density}.}
    \label{fig:teaser-graph}
    \vspace*{-1.em}
\end{figure}

Our analysis suggests the performance bottleneck may stem from the systematic elimination of valuable low-probability exploratory tokens. We term these tokens \textbf{Reasoning Sparks}; they include words like ``wait'', ``however'', or ``perhaps'', which often serve as logical connectives or expressions of uncertainty that naturally initiate diverse reasoning pathways (Figure~\ref{fig:a}). As the aggregated violin plots in Figure~\ref{fig:c} show, standard GRPO training suppresses the low-probability sampling of these important exploratory tokens, causing the suppression of \textit{Reasoning Sparks}. Furthermore, we find that indiscriminately boosting randomness amplifies the low-probability sampling of irrelevant tokens (e.g., ``cost'', ``fine''), which are semantically out of context for the mathematical reasoning task. We refer to the low-probability appearance of these irrelevant tokens as noise. This amplification leads to an even faster performance collapse than the baseline, as shown in Figure~\ref{fig:b}.

These findings present a central challenge: \textbf{a successful exploration strategy should protect valuable \textit{reasoning sparks} without simultaneously amplifying the destructive effects of irrelevant noise.} To address this challenge, we introduce Low-probability Regularization (Lp-Reg). The primary goal of Lp-Reg is to preserve valuable low-probability tokens via regularization. To avoid amplifying noise, the method leverages a key observation: within the low-probability range, meaningful exploratory tokens (like ``wait'') consistently exhibit a higher average probability than irrelevant noise (like ``cost'') in the immediate next-token prediction. Based on this statistical distinction, Lp-Reg first discards low-probability tokens presumed to be noise using a probability threshold. It then redistributes the probability mass from these discarded tokens among the remaining candidates. This process constructs a less-noisy proxy distribution where valuable low-probability tokens are preserved and their relative probabilities amplified. Finally, Lp-Reg penalizes the deviation of the original policy from this proxy using a forward KL divergence, which selectively protects the low-probability tokens that were preserved in the less-noisy proxy distribution.

Our experimental evaluation demonstrates the effectiveness of Lp-Reg. Our method enables stable on-policy training for around $3,000$ steps over $81,204$ GPU-hours, a regime where many entropy-control methods have collapsed, resulting in better performance. On five widely used math benchmarks, this results in a $60.17\%$ average accuracy on Qwen3-14B-Base, improving upon prior methods by $2.66\%$. Our contributions are summarized as follows:

\begin{itemize}
    \item In contrast to prior work focusing on overall policy entropy, we identify the suppression of \textit{reasoning sparks} as a key issue and provide evidence that their preservation is crucial for sustained performance.
    \item We introduce Low-probability Regularization (Lp-Reg), a method that creates a more stable exploratory environment by filtering out presumed meaningless noise to protect the remaining low-probability tokens.
    \item We demonstrate through extensive experiments that Lp-Reg achieves state-of-the-art performance, while also enabling stable on-policy continuous scaling over extended periods where baselines collapse.
    \item We provide a comprehensive analysis showing that our approach of filtering presumed meaningless noise yields superior results compared to indiscriminate entropy-control methods.
\end{itemize}
\section{Related Work}

\paragraph{Reinforcement learning for LLMs} Recently, reinforcement learning has become the dominant framework for enhancing the reasoning abilities of large language models (LLMs)~\citep{openai2024openaio1card, deepseekai2025deepseekr1incentivizingreasoningcapability}. By leveraging automatic checkers or symbolic verification, reinforcement learning with verifiable rewards (RLVR) achieved further breakthroughs in improving the reasoning capability of LLMs~\citep{shao2024deepseekmathpushinglimitsmathematical, yang2025qwen3technicalreport, kimiteam2025kimik15scalingreinforcement}. Based on RLVR and GRPO~\citep{shao2024deepseekmathpushinglimitsmathematical}, subsequent methods such as DAPO~\cite{yu2025dapoopensourcellmreinforcement}, VAPO~\citep{yue2025vapoefficientreliablereinforcement}, and other policy optimization variants~\citep{zhao2025geometricmeanpolicyoptimization, cui2025entropymechanismreinforcementlearning, zheng2025groupsequencepolicyoptimization} have been proposed to improve the stability, efficiency, and scalability of RL for reasoning models.

\paragraph{Entropy collapse in RL training} A recurring difficulty in training reasoning models with RL is the rapid collapse of policy entropy during the early stages of training. This phenomenon, which reflects excessive exploitation and insufficient exploration, has been widely recognized as a bottleneck for scaling RL in reasoning models. To mitigate collapse, researchers have explored several directions, including selectively regularizing updates at high-entropy “forking” tokens~\citep{wang20258020rulehighentropyminority}, amplifying advantages at exploratory positions~\citep{cheng2025reasoningexplorationentropyperspective}, modifying clipping strategies~\citep{yu2025dapoopensourcellmreinforcement, zhao2025geometricmeanpolicyoptimization, cui2025entropymechanismreinforcementlearning, zheng2025groupsequencepolicyoptimization}, or doing weight clipping~\citep{minimax2025minimaxm1scalingtesttimecompute, su2025klearreasoneradvancingreasoningcapability}. While these methods primarily operate by monitoring policy entropy, which is correlational rather than causal to exploration, our analysis delves directly into the next-token prediction distribution. This allows for a more semantically grounded and causally-informed investigation of the probabilities of individual candidates and their role in exploration dynamics.

\paragraph{Intrinsic confidence of LLMs} As the capabilities of Large Language Models (LLMs) have rapidly advanced, they have demonstrated an increasingly strong and reliable sense of intrinsic confidence~\citep{saurav2022language, loka2024confidence, amir2025confidence}. Research investigates how these intrinsic confidence signals, often reflected in the next-token prediction distribution, can guide complex reasoning and exploration~\citep{amirhosein2025guided, xuezhi2024chainofthought, xuandong2025learning}. Studies have shown that tokens with higher relative probabilities in the next-token prediction are often more contextually appropriate than their lower-probability counterparts~\citep{nguyen2025turningheatminpsampling, xu2025adaptiveterminationmultiroundparallel, fu2025deepthinkconfidence}. Building on this, some work has explored entropy minimization, which sharpens the model's confidence distribution. This approach can improve inference performance by encouraging the model to commit to consistent and high-confidence solution paths~\citep{gao2025oneshotentropyminimization, agarwal2025unreasonableeffectivenessentropyminimization}. Our work builds upon a similar insight, leveraging the model's intrinsic confidence to distinguish between valuable \textit{reasoning sparks} and irrelevant noise within this low-probability range.

\section{Preliminaries}
\subsection{Reinforcement Learning with Verifiable Rewards}
Reinforcement learning (RL) has played a critical role in LLMs \citep{murphy2024reinforcement}. Formally,
\begin{equation}
\mathcal{J}_{\text{RL}}(\boldsymbol{\theta}) = \mathbb{E}_{(q,a) \sim D, o \sim \pi_{\theta}(\cdot\mid q)} \big[ r(o,a) \big],
\label{eq:rl}
\end{equation}
where $r(o,a)$ denotes the reward assigned to an output $o$ given a reference answer $a$. In reinforcement learning with verifiable rewards (RLVR), this reward is computed through rule-based functions, such as Math-Verify\footnote{\url{https://github.com/huggingface/Math-Verify}}. Recent studies have demonstrated that large-scale RLVR encourages models to perform more deliberative reasoning by producing extended chains of thought prior to the final prediction, thereby substantially improving their capacity to solve complex problems \citep{deepseekai2025deepseekr1incentivizingreasoningcapability}. In practice, Eq.~\ref{eq:rl} is typically optimized using policy gradient methods, such as Proximal Policy Optimization (PPO) \citep{schulman2017proximal} and Group Relative Policy Optimization (GRPO) \citep{shao2024deepseekmath}.

\subsection{Group-Relative Policy Optimization}
GRPO is a representative actor-only policy gradient method for optimizing LLMs. It directly estimates the advantage of each token by leveraging multiple samples drawn from the same prompt. Formally, the advantage is defined as
\begin{equation}
    \begin{aligned}
        A_{i,t} &= \frac{R(o_i) - \mathrm{mean}(\mathcal{G})}{\mathrm{std}(\mathcal{G})},
    \end{aligned} 
\end{equation}
where $\{o_1, \ldots, o_G\}$ are independent outputs sampled from the same prompt, with group size $G$, $\mathcal{G} = \{R(o_1), \ldots, R(o_G)\}$ denotes their associated rewards, and $R(o_i)$ is the reward of output $o_i$. In this formulation, $A_{i,t}$ represents the advantage of the $t$-th token in $o_i$. The policy is then optimized on the basis of these advantages using the PPO surrogate objective:
\begin{equation}
    \begin{aligned}
        \mathcal{J}_{\mathrm{GRPO}}(\boldsymbol{\theta}) = & \mathbb{E}_{\,(q,a)\sim D,\,\{o_i\}_{i=1}^G \sim \pi_{\boldsymbol{\theta}_{\mathrm{old}}}(\cdot\mid q)} \\
        & \frac{1}{G}\sum_{i=1}^G \frac{1}{\lvert o_i\rvert}
           \sum_{t=1}^{\lvert o_i\rvert}
            \Bigl[
             \min\bigl[
                 r_{i,t}\, A_{i,t},\, \mathrm{clip}\bigl(r_{i,t},1-\epsilon,1+\epsilon\bigr)\, A_{i,t}\bigr] - \beta\,D_{\mathrm{KL}}\left(\pi_{\boldsymbol{\theta}}\,\|\,\pi_{\mathrm{ref}}\right) \Bigl],
    \end{aligned}
\end{equation}
where $\beta$ controls the strength of KL regularization between the current policy $\pi_{\theta}$ and the reference policy $\pi_{\mathrm{ref}}$. The probability ratio
\begin{equation}
    \begin{aligned}
        r_{i,t}
        = \frac{\pi_{\theta}\bigl(o_{i,t} \mid q, o_{i,<t}\bigr)}
        {\pi_{\theta_{\mathrm{old}}}\bigl(o_{i,t} \mid q, o_{i,<t}\bigr)}
    \end{aligned}
\end{equation}
serves as the importance sampling weight for off-policy training, where $\pi_{\theta_{\mathrm{old}}}$ denotes the behavior policy. The hyperparameter $\epsilon$ specifies the clipping ratio, which constrains the updated policy from deviating excessively from the behavior policy, thereby ensuring stability during optimization.
\section{Low-probability Regularization}

To address the premature elimination of valuable \textit{reasoning sparks}, we propose a regularization method termed \textbf{Low-probability Regularization (Lp-Reg)}. This method is designed to be integrated into policy gradient algorithms to create a more stable exploratory environment. The central idea is to leverage the model's own predictive distribution to construct a less-noisy proxy for regularization, preserving low-probability tokens.

\subsection{Proxy Distribution $\pi_{\text{proxy}}$}\label{sec:Proxy Distribution}
The foundation of Lp-Reg is the construction of a proxy distribution, which represents a filtered variant of the current policy $\pi_{\boldsymbol{\theta}}$. It is constructed in two steps:

\begin{figure}[!h]
    \centering 
    \includegraphics[width=1.0\textwidth]{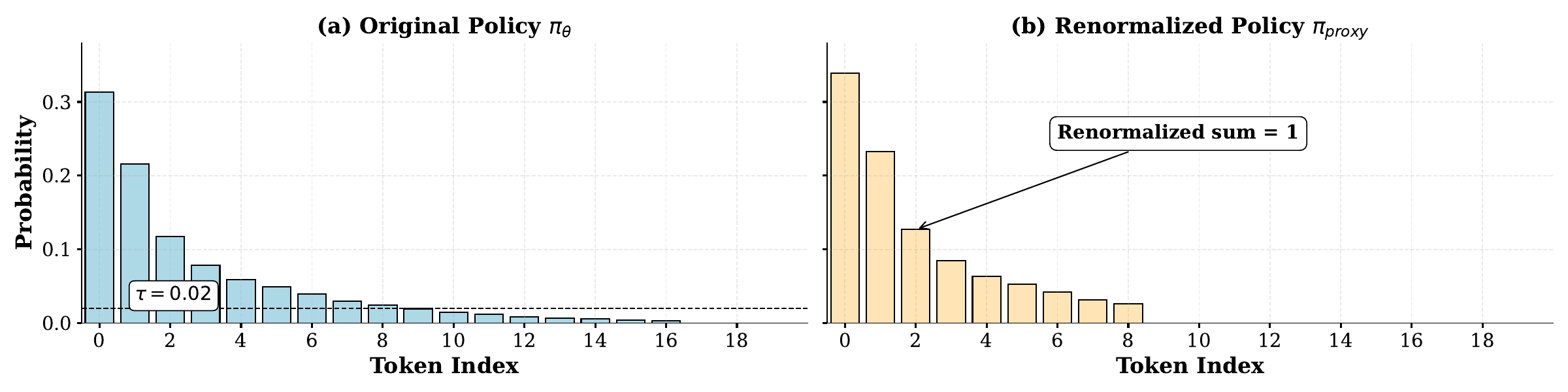}
    \caption{An example of probability renormalization. $\pi_{\text{proxy}}$ assigns zero probability to tokens with $\pi_{\boldsymbol{\theta}} \leq \tau$ and renormalizes the probability mass to tokens with $\pi_{\boldsymbol{\theta}} > \tau$.}
    \label{fig:policy_renorm}
\end{figure}

\begin{enumerate}[leftmargin=2em]
    \item \textbf{Filtering Noise Tokens:} We first filter out a set of low-confidence tokens, which are presumed to be noise, defined as those whose probability $\pi_{\boldsymbol{\theta}}(o|\cdot)$ is under a threshold $\tau$. This threshold controls the filtering strategy, for which we explore two primary choices:
    \begin{itemize}[leftmargin=2em]
        \item \textbf{Fixed threshold:} A simple approach where $\tau$ is a constant hyperparameter, e.g., $\tau=0.02$.
        \item \textbf{Min-p threshold:} Following~\citep{nguyen2025turningheatminpsampling}, $\tau$ is defined relative to the peak probability: $\tau = \kappa \cdot \max_{o' \in V} \pi_{\boldsymbol{\theta}}(o'|\cdot)$, where $\kappa\in(0,1)$ is a hyperparameter. This makes the filtering adaptive to the distribution's sharpness.
    \end{itemize}
    Our primary experiments employ the min-p strategy for its adaptiveness, though fixed thresholds are also shown to be effective in our ablation studies.

    \item \textbf{Probability Renormalization:} As shown in Figure~\ref{fig:policy_renorm}, the proxy distribution $\pi_{\text{proxy}}$ assigns zero probability to tokens filtered out in the previous step and renormalizes the probability mass across the remaining tokens:
    \begin{equation}
    \pi_{\text{proxy}}(o|\cdot) = 
    \begin{cases} 
        \frac{\pi_{\boldsymbol{\theta}}(o|\cdot)}{\sum_{o' \text{ s.t. } \pi_{\boldsymbol{\theta}}(o'|\cdot) > \tau} \pi_{\boldsymbol{\theta}}(o'|\cdot)} & \text{if } \pi_{\boldsymbol{\theta}}(o|\cdot) > \tau \\
        0 & \text{otherwise}
    \end{cases}.
    \label{eq:proxy_dist}
    \end{equation}
    This process effectively treats tokens with low relative probabilities as potential noise, while preserving all others to form a high-confidence reference.
\end{enumerate}

\subsection{Low-probability Regularization Objective}
The Low-probability Regularization (Lp-Reg) penalty is integrated into the GRPO framework as a selective regularization term. The final objective function is:
\begin{equation}
\begin{aligned}
    J_{\text{Lp-Reg}}(\boldsymbol{\theta}) = & \mathbb{E}_{\mathcal{B}\sim\mathcal{D}, (q, a) \sim \mathcal{B}, \{o_i\}_{i=1}^G\sim \pi_{\boldsymbol{\theta}_{\text{old}}}(\cdot|q)} 
    \Bigg[ \frac{1}{\sum_{i=1}^G|o_i|} \sum_{i=1}^G \sum_{t=1}^{|o_i|} \bigg[ \mathrm{clip}(r_{i, t}(\boldsymbol{\theta}), 0, U) \cdot A_{i, t} \\
    & - \beta \cdot \mathbb{I}\bigg[ \underbrace{\pi_{\boldsymbol{\theta}}(o_{i, t} |q, o_{i, <t}) < \delta_{\rho}^{\mathcal{B}}}_{\text{low probability}} \land \underbrace{\pi_{\text{proxy}}(o_{i, t} |q, o_{i, <t}) > 0}_{\text{not noise}} \land \underbrace{A_{i,t}<0}_{\text{neg. sample}} \bigg] \\
    & \qquad \cdot \mathcal{D}_\mathrm{KL}\left( \pi_{\text{proxy}}(\cdot |q, o_{i, <t}) \Vert \pi_{\boldsymbol{\theta}}(\cdot |q, o_{i, <t}) \right) \bigg] \Bigg]
\end{aligned}
\label{eq:lp_reg_objective}
\end{equation}
The first term is the policy gradient objective from GRPO. We modify its clipping by removing the lower bound to avoid clipping low-probability exploratory actions and adding a large upper bound $U$ for numerical stability.

The second term is the Lp-Reg penalty, designed to protect \textit{reasoning sparks}. It is activated by the indicator function $\I[\cdot]$ only for tokens that satisfy three conditions simultaneously: first, their sampling probability $\pi_{\boldsymbol{\theta}}$ is below a dynamic low-percentile threshold $\delta_{\rho}^{\mathcal{B}}$, which is calculated as the lowest $\rho$-th percentile of the sampling probabilities of all tokens within the current training batch $\mathcal{B}$; second, their probability in the proxy distribution $\pi_{\text{proxy}}$ is greater than zero, means $o_{i,t}$ is not a noise token; and third, the token receives a negative advantage signal ($A_{i,t} < 0$). This final condition ensures the regularization applies exclusively to tokens receiving a negative learning signal, preventing their potential over-penalization while leaving updates from positive experiences unaffected. We ablate all three conditions in Section~\ref{sec:Ablation Study}.

We use the forward KL divergence, $\mathcal{D}_\mathrm{KL}(\pi_{\text{proxy}} \Vert \pi_{\boldsymbol{\theta}})$ as the regularization function. It imposes a significant penalty when $\pi_{\boldsymbol{\theta}}(o|\cdot)$ approaches zero for a token $o$ with non-zero probability in $\pi_{\text{proxy}}$, providing a targeted penalty against token elimination without forcing the policy to strictly match the heuristic proxy distribution.
\section{Experiments}

\subsection{Experimental Setup}

\subsubsection{Baselines}
We compare Lp-Reg against a suite of strong baselines, including a foundational algorithm and several state-of-the-art methods designed to enhance exploration through entropy control. Our primary baseline is \textbf{GRPO}~\citep{shao2024deepseekmathpushinglimitsmathematical}, a value-free policy optimization algorithm that employs group-relative advantage estimation, making it a common choice for RLVR. To represent classical entropy regularization methods, we implement \textbf{GRPO + Entropy Loss}, which directly incorporates the principles of Maximum Entropy RL by adding a policy entropy bonus to the GRPO objective function. We also compare against several advanced methods: \textbf{Clip-Higher}~\citep{yu2025dapoopensourcellmreinforcement}, a core component of DAPO that encourages higher entropy by using an asymmetric clipping range in the PPO objective; \textbf{Selective High-Entropy Training (80/20)}~\citep{wang20258020rulehighentropyminority}, a method that restricts policy gradient updates to only the top 20\% of tokens with the highest generation entropy; \textbf{KL-Cov}~\citep{cui2025entropymechanismreinforcementlearning}, which prevents entropy collapse by applying a selective KL-divergence penalty to tokens with the highest covariance between their log probabilities and advantages; and \textbf{GSPO}~\citep{zheng2025groupsequencepolicyoptimization}, which modifies the clipping mechanism to operate at the sequence level to promote higher training entropy.

\subsubsection{Training Settings}
All experiments are conducted within the \texttt{verl}~\citep{sheng2024hybridflow} framework to ensure a standardized and fair comparison. To mitigate unfair comparisons arising from different convergence speeds during the early stages of training, we trained models until their performance saturated. This approach ensures a more stable and equitable evaluation point. Consequently, our comparisons are based on approximately $1,000$ training steps for the Qwen3-14B-Base model and $800$ for the Qwen2.5-32B model. Each training requires about $8,000$ GPU hours on 32 NVIDIA H20 GPUs for the 14B model and $16,000$ GPU hours on 64 NVIDIA H20 GPUs for the 32B model. However, for models that experienced a training collapse, which we define as a performance drop greater than $10\%$ in accuracy, we implemented early stopping to conserve computational resources. To assess whether low-probability tokens sustain exploration in RLVR, we further trained the Qwen2.5-32B model for $3,000$ steps over $81,204$ GPU-hours with our Lp-Reg and evaluated its training stability.

For the reinforcement learning from verifier rewards (RLVR) phase, models are trained on the Dapo-Math-17K~\citep{yu2025dapoopensourcellmreinforcement} dataset with a maximum response length of $8,192$ tokens. We use a global batch size of 256. For off-policy methods, we use a mini-batch size of 32, resulting in 8 gradient updates per rollout. To ensure a fair comparison, a ``step'' in our experimental results consistently refers to a single rollout for all methods. Consequently, each reported step for off-policy training corresponds to 8 gradient updates. A constant learning rate of $1 \times 10^{-6}$ is applied without a warmup schedule. We set the group number as $8$ for all GRPO-based methods. To ensure numerical stability, we set the policy gradient's clipping by setting the upper bound of the importance sampling ratio to $U=10$. For our proposed Lp-Reg, which uses the min-p threshold, we set the probability percentile threshold $\rho$ to $0.5\%$ for Qwen2.5-32B-Base and $1\%$ for Qwen3-14B-Base, the KL regularization coefficient $\beta$ to $1.0$, and the min-p ratio $\kappa$ to $0.02$. The proxy distribution, $\pi_{\text{proxy}}$, is constructed from the data-generating policy ($\pi_{\theta_{\text{old}}}$ in the off-policy setting and the current policy $\pi_{\boldsymbol{\theta}}$ in the on-policy setting). For all baseline methods, we adopt the hyperparameters specified in their original public implementations to ensure a faithful reproduction. Specifically for the GRPO + Entropy Loss baseline, we set the entropy coefficient to $0.002$ within the \texttt{verl} framework.

\subsubsection{Evaluation}
For evaluation, we assess model performance across five diverse mathematical reasoning benchmarks: AIME24~\citep{aime}, AIME25~\citep{aime}, MATH-500~\citep{hendrycks2021measuring}, OlympiadBench~\citep{he-etal-2024-olympiadbench}, and Minerva Math~\citep{lewkowycz2022solving}. Following \citep{cui2025entropymechanismreinforcementlearning}, we employ distinct decoding strategies based on the benchmark. For AIME24 and AIME25, which have smaller test sets, we use sampled decoding with a temperature of $0.6$ and generate $16$ independent responses per problem to obtain a robust performance estimate. For the remaining benchmarks, including MATH-500, OlympiadBench, and Minerva, we utilize greedy decoding to evaluate performance.

\subsection{Results}

\begin{figure}[!h]
    \centering 
    \includegraphics[width=1.0\textwidth]{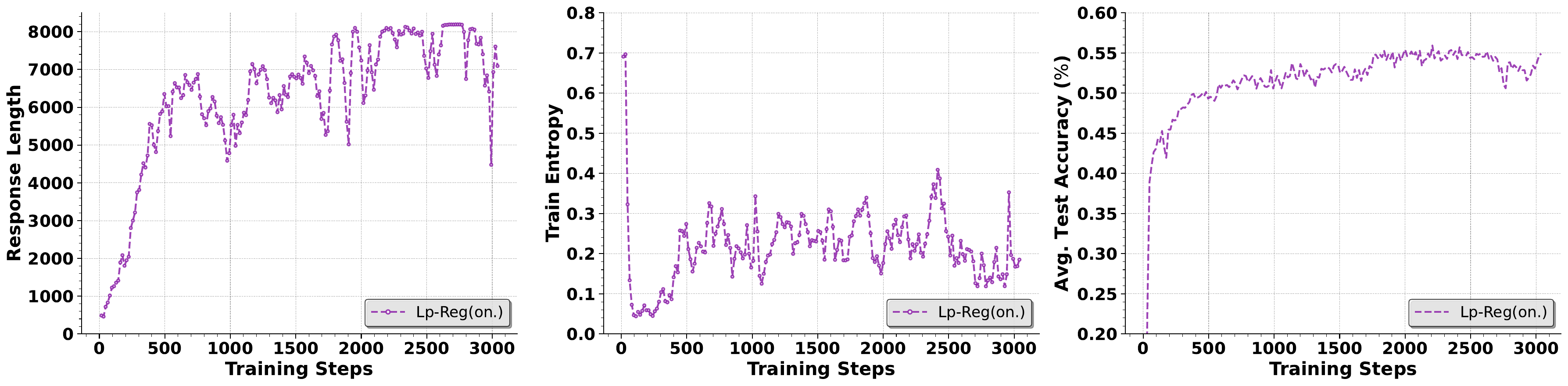}
    \vspace{-20pt}
    \caption{\textbf{Continuous scaling over $3,000$ training steps, totaling $81,204$ GPU-hours}, for Lp-Reg (on-policy) on the Qwen2.5-32B-Base model.}
    \label{fig:32b_scaling_combined_metrics}
\vspace{-5pt}
\end{figure}

\begin{figure}[!h]
    \centering 
    \vspace{-10pt}
    \includegraphics[width=0.75\textwidth]{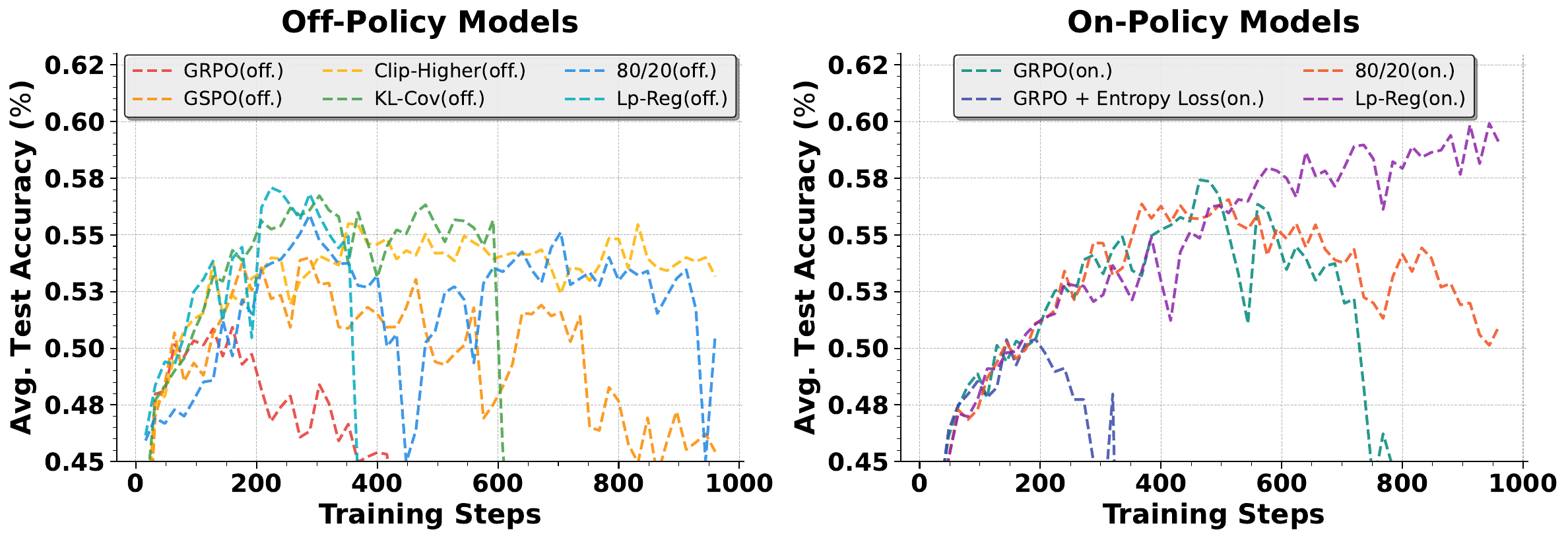}
    \vspace{-10pt}
    \caption{Training dynamics on the Qwen3-14B-Base model. On-policy training exhibits better training stability and testing performance compared to off-policy training.}
    \label{fig:14b_offpolicy_onpolicy_comparison}
    \vspace{-5pt}
\end{figure}

As shown in Figure~\ref{fig:32b_scaling_combined_metrics}, Lp-Reg enables a continuous reinforcement learning scaling for $3,000$ training steps, totaling $81,204$ GPU-hours on the Qwen2.5-32B-Base model. Furthermore, Figure~\ref{fig:14b_offpolicy_onpolicy_comparison} and Table~\ref{tab:main-results} reveal that Lp-Reg achieves state-of-the-art performance across five challenging mathematical reasoning benchmarks on both 14B and 32B model scales. On the Qwen3-14B model, on-policy Lp-Reg sets a new benchmark with an average accuracy of $60.17\%$, surpassing the next best method, 80/20, by $2.66\%$. Notably, Lp-Reg’s advantage is more pronounced on the newer Qwen3-14B base model compared to the older Qwen2.5-32B in the first $1,000$ training steps. We hypothesize that as base models improve, their capacity for nuanced, low-probability reasoning increases, creating a richer substrate for the emergence of valuable \textit{reasoning sparks}, which Lp-Reg can then effectively protect and leverage.

Our experiments consistently show the superiority of on-policy training over off-policy methods across 14B and 32B scales in Figure~\ref{fig:14b_offpolicy_onpolicy_comparison}. This is due to the inherent stability of on-policy updates, which avoid distribution shifts caused by mismatched data-sampling and training policies. Off-policy methods, such as Clip-Higher, often rely on importance sampling clipping, leading to instability. While competitive on Qwen2.5-32B, Clip-Higher's performance drops on Qwen3-14B, highlighting its fragility. In contrast, Lp-Reg’s self-contained, policy-intrinsic regularization ensures its effectiveness in both on-policy and off-policy settings, unlike competing methods that are heavily reliant on off-policy importance sampling.

\begin{figure}[!h]
    \centering 
    \vspace{-5pt}
    \includegraphics[width=1.0\textwidth]{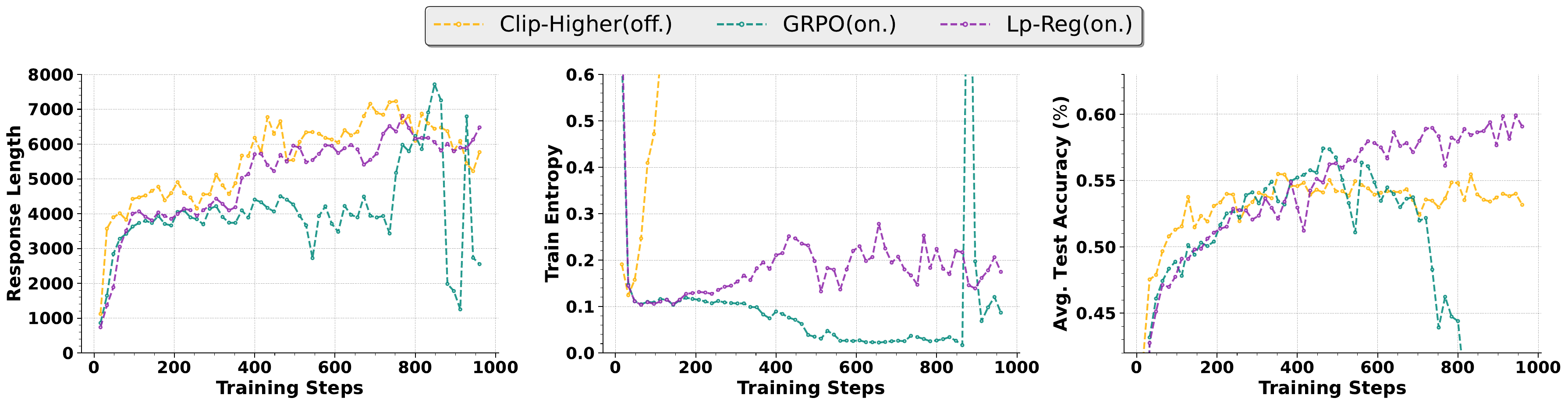}
    \vspace{-15pt}
    \caption{Training dynamics on the Qwen3-14B-Base model. To best illustrate the performance differences, we compare the top-performing methods. Lp-Reg demonstrates more stable performance throughout training.}
    \label{fig:14b_combined_metrics}
    \vspace{-5pt}
\vspace{-10pt}
\end{figure}

Beyond raw performance, Lp-Reg demonstrates a distinct entropy signature indicative of a healthy exploration-exploitation balance. As shown in Figure~\ref{fig:14b_combined_metrics}, methods like Clip-Higher induce a continuous, often artificial increase in policy entropy. Lp-Reg, however, facilitates a dynamic, multi-phase entropy trajectory: entropy initially decreases as the model learns core reasoning patterns, then gradually increases to foster exploration as performance improves, and finally stabilizes within a healthy range as accuracy converges. This adaptive behavior stems from our confidence-aware regularization, which selectively protects \textit{reasoning sparks} without amplifying low-probability out-of-context irrelevant noise.

\begin{table}[!h]
\centering
\scalebox{0.85}{
\begin{tabular}{l|ccccc|c}
\toprule
Method & AIME24 & AIME25 & Math-500 & Minerva & Olympiad Bench & Avg. \\ 
\midrule[\heavyrulewidth]
\multicolumn{7}{c}{\textbf{Qwen2.5-32B-Base} ($800$ training steps)} \\
\midrule
\cellcolor{offpolicycolor} GRPO~\citep{shao2024deepseekmathpushinglimitsmathematical} (off.) & 30.63 & 22.29 & 88.00 & 41.18 & 54.37 & 47.29 \\
\cellcolor{offpolicycolor} GSPO~\citep{zheng2025groupsequencepolicyoptimization} (off.) & 33.33 & 22.29 & 87.60 & \textbf{48.53} & 55.56 & 49.46 \\
\cellcolor{offpolicycolor} Clip-Higher~\citep{yu2025dapoopensourcellmreinforcement} (off.) & \textbf{38.33} & \textbf{29.79} & 87.60 & 45.22 & 56.44 & 51.48 \\
\cellcolor{offpolicycolor} KL-Cov~\citep{cui2025entropymechanismreinforcementlearning} (off.) & 35.62 & 27.50 & 87.40 & 44.49 & 55.11 & 50.02 \\
\cellcolor{offpolicycolor} 80/20~\citep{wang20258020rulehighentropyminority} (off.) & 38.12 & 28.75 & 87.00 & 45.22 & 58.37 & 51.49 \\
\cellcolor{offpolicycolor} Lp-Reg (off.) & 37.71 & 24.58 & \textbf{90.20} & 40.81 & 59.70 & 50.60 \\
\cellcolor{onpolicycolor} GRPO~\citep{shao2024deepseekmathpushinglimitsmathematical} (on.) & 28.54 & 22.50 & 86.60 & 44.85 & 60.30 & 48.56 \\
\cellcolor{onpolicycolor} GRPO + Entropy Loss (on.) & 3.75 & 1.88 & 60.80 & 27.94 & 22.22 & 23.32 \\
\cellcolor{onpolicycolor} 80/20~\citep{wang20258020rulehighentropyminority}(on.) & 32.50 & 28.54 & 89.40 & 45.59 & 57.63 & 50.73 \\
\cellcolor{onpolicycolor} Lp-Reg (on.) & 38.12 & 27.08 & 90.00 & 46.32 & \textbf{61.19} & \textbf{52.54} \\
\midrule[\heavyrulewidth]
\multicolumn{7}{c}{\textbf{Qwen3-14B-Base} ($1,000$ training steps)} \\
\midrule
\cellcolor{offpolicycolor} GRPO~\citep{shao2024deepseekmathpushinglimitsmathematical} (off.) & 34.38 & 27.08 & 89.20 & 49.26 & 55.70 & 51.13 \\
\cellcolor{offpolicycolor} GSPO~\citep{zheng2025groupsequencepolicyoptimization} (off.) & 41.46 & 34.58 & 88.60 & \textbf{50.74} & 59.85 & 55.05 \\
\cellcolor{offpolicycolor} Clip-Higher~\citep{yu2025dapoopensourcellmreinforcement} (off.) & 41.67 & 32.71 & \textbf{95.00} & 47.43 & 64.00 & 56.16 \\
\cellcolor{offpolicycolor} KL-Cov~\citep{cui2025entropymechanismreinforcementlearning} (off.) & 49.17 & 34.79 & 93.00 & 47.43 & 62.07 & 57.29 \\
\cellcolor{offpolicycolor} 80/20~\citep{wang20258020rulehighentropyminority} (off.) & 43.96 & 34.58 & 91.80 & 48.16 & 60.89 & 55.88 \\
\cellcolor{offpolicycolor} Lp-Reg (off.) & 46.25 & 34.17 & 92.40 & 48.16 & 64.44 & 57.08 \\
\cellcolor{onpolicycolor} GRPO~\citep{shao2024deepseekmathpushinglimitsmathematical} (on.) & 46.04 & 34.38 & 93.00 & 48.53 & 65.19 & 57.43 \\
\cellcolor{onpolicycolor} GRPO + Entropy Loss (on.) & 37.29 & 25.21 & 88.20 & 46.32 & 54.96 & 50.40 \\
\cellcolor{onpolicycolor} 80/20~\citep{wang20258020rulehighentropyminority} (on.) & 47.29 & 32.50 & 91.60 & 50.37 & 65.78 & 57.51 \\
\cellcolor{onpolicycolor} Lp-Reg (on.) & \textbf{50.83} & \textbf{37.92} & 94.40 & 49.26 & \textbf{68.44} & \textbf{60.17} \\
\bottomrule
\end{tabular}}
\caption{Main results on five mathematical reasoning benchmarks across two model scales. On-policy (on.) and off-policy (off.) training methods are highlighted with distinct colors. We report the best score among the total training for each method. }
\label{tab:main-results}
\end{table}

\begin{figure}[!h]
    \centering 
    \includegraphics[width=1.0\textwidth]{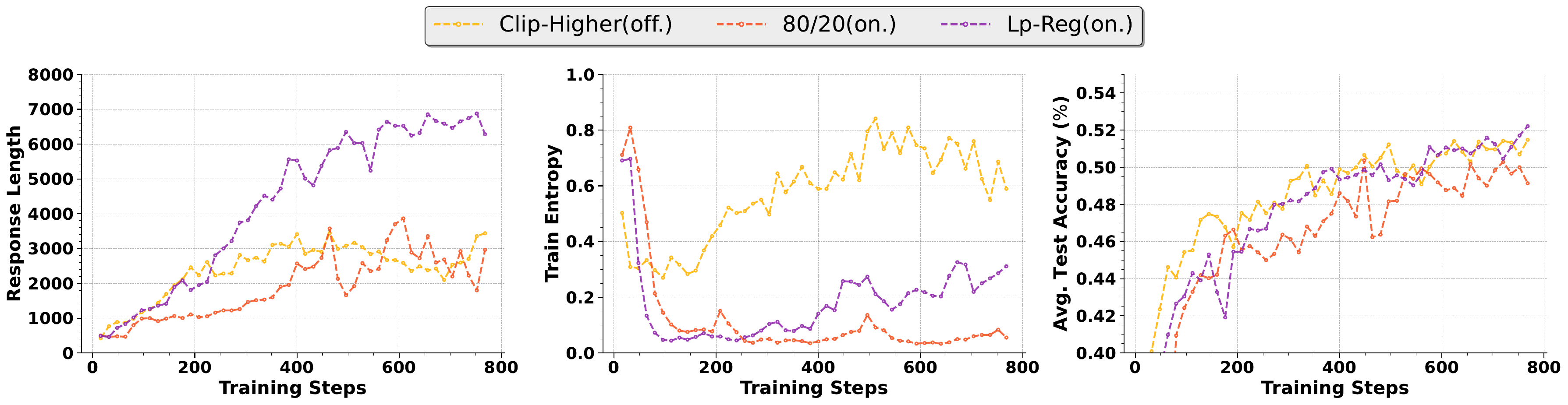}
    \vspace{-20pt}
    \caption{Training dynamics on the Qwen2.5-32B-Base model. To best illustrate the performance differences, we compare the top-performing methods.}
    \label{fig:32b-main-results}
\vspace{-5pt}
\end{figure}

\subsection{Ablation Study} \label{sec:Ablation Study}

We conduct a series of ablation studies to analyze the core components of Lp-Reg and validate our key design choices for regularization in Equation~\ref{eq:lp_reg_objective}, including low-probability regularization, noise filtering, negative sample regularization, and forward KL.

\begin{figure}[!t]
    \centering 
    \includegraphics[width=1\textwidth]{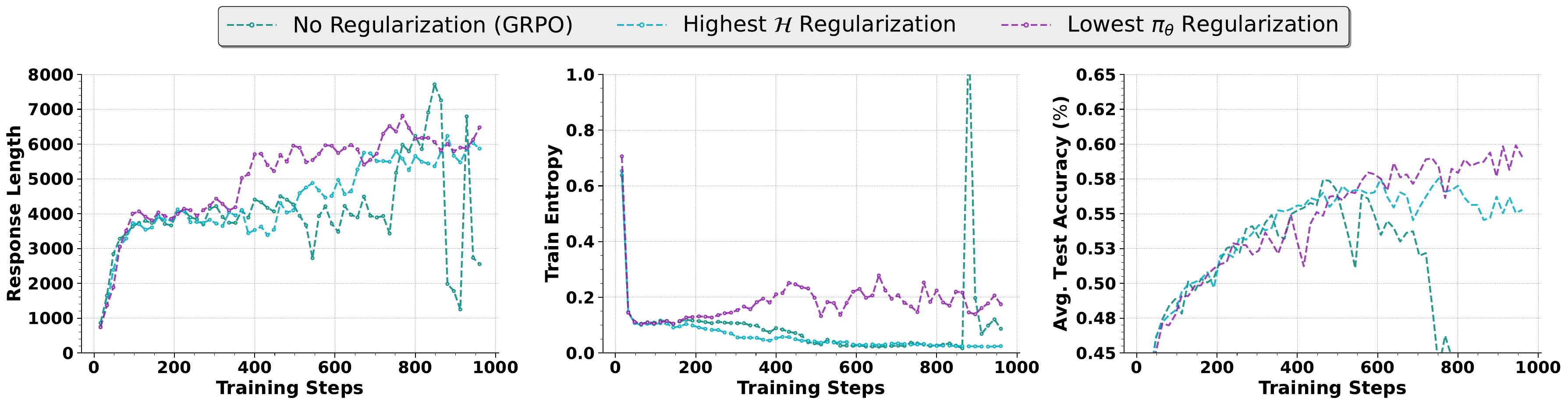}
    \vspace{-20pt}
    \caption{Ablation study comparing \textbf{low-probability token regularization versus high-entropy token regularization} for Lp-Reg (on-policy) on the Qwen3-14B-Base model.}
    \label{fig:14b_ablation_extra_combined_metrics}
\vspace{-5pt}
\end{figure}

\paragraph{Importance of Low-Probability Token}
To verify that targeting low-probability tokens is superior to the conventional wisdom of targeting high entropy, we conduct a comparison between the high-entropy token regularization (w/ highest $\mathcal{H}$ regularization) and the low-probability regularization (w/ lowest $\pi_{\boldsymbol{\theta}}$ regularization, vanilla Lp-Reg). Instead of applying Lp-Reg to the lowest $1\%$ probability tokens, we apply an identical regularization mechanism to the tokens with the highest $1\%$ entropy. As shown in Figure~\ref{fig:14b_ablation_extra_combined_metrics}, this approach not only fails to improve performance but also fails to sustain high entropy, which collapses after an initial spike. This result reinforces our claim from the Introduction: high entropy is a poor proxy for valuable exploration. As our analysis in Section~\ref{sec:Qualitative Analysis of Targeted Tokens} further corroborates, high-entropy tokens are often common function words or formatting characters, not the meaningful, low-probability exploratory tokens we term \textit{reasoning sparks}. Regularizing them pollutes the learning signal without protecting the structured, low-probability reasoning paths necessary for progress.

\begin{figure}[!t]
    \centering
    \includegraphics[width=1.0\linewidth]{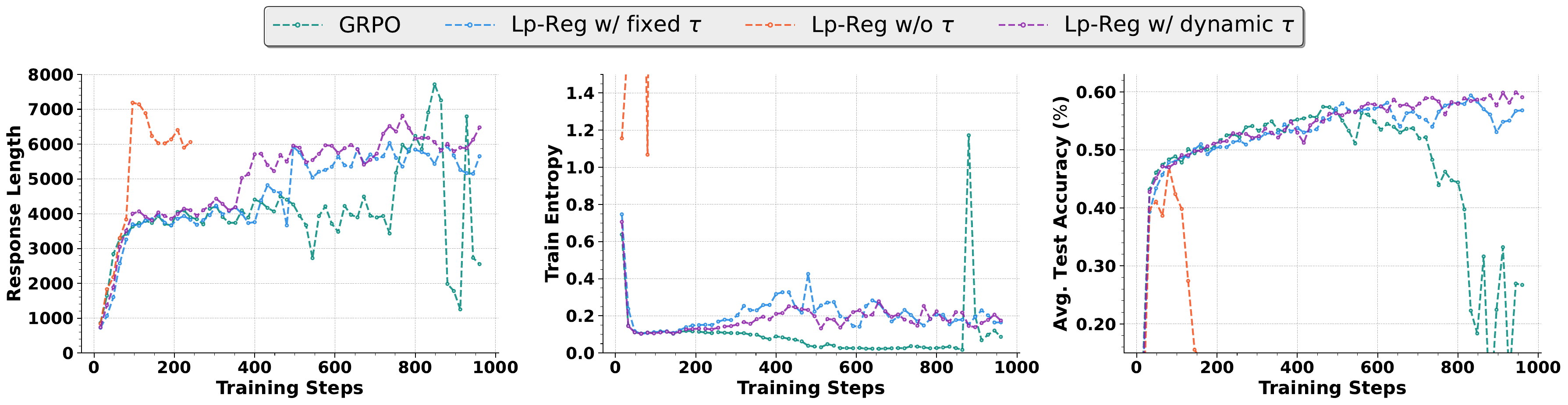}
    \vspace{-20pt}
    \caption{Ablation studies for \textbf{noise filtering} of Lp-Reg (on-policy) on the Qwen3-14B-Base model. The results confirm that targeting our noise filtering threshold $\tau$ is critical for stable performance. The adaptiveness of the min-p threshold is also shown to be beneficial over a fixed one.}
    \label{fig:14b_ablation_main_combined_metrics}
\vspace{-5pt}
\end{figure}

\paragraph{Importance of Noise Filtering.}
Lp-Reg only protects tokens deemed meaningful by the proxy distribution ($\pi_{\text{proxy}} > 0$). To test this, we remove the filter and fork all tokens below the noise threshold $\tau$ from contributing to gradient updating (Lp-Reg w/o $\tau$). Figure~\ref{fig:14b_ablation_main_combined_metrics} shows that this leads to a catastrophic performance collapse and entropy explosion. This confirms that filtering is critical to ignore the extreme tail of the distribution, which consists of irrelevant noise that destabilizes training if regularized. 
We further conduct a comparison between the dynamic min-p noise threshold (Lp-Reg w/ dynamic $\tau$) and the fixed noise threshold (Lp-Reg w/ fixed $\tau$) in Section~\ref{sec:Proxy Distribution}. As shown in Figure~\ref{fig:14b_ablation_main_combined_metrics}, the fixed threshold underperforms compared to the dynamic threshold, which we adopt as the default. However, it still significantly surpasses the standard GRPO. This indicates that while the core filtering principle is effective, the dynamic nature of min-p provides a more robust estimate of the model's confidence across different contexts, better preserving genuine \textit{reasoning sparks}.

\begin{figure}[!t]
    \centering 
    \includegraphics[width=1\textwidth]{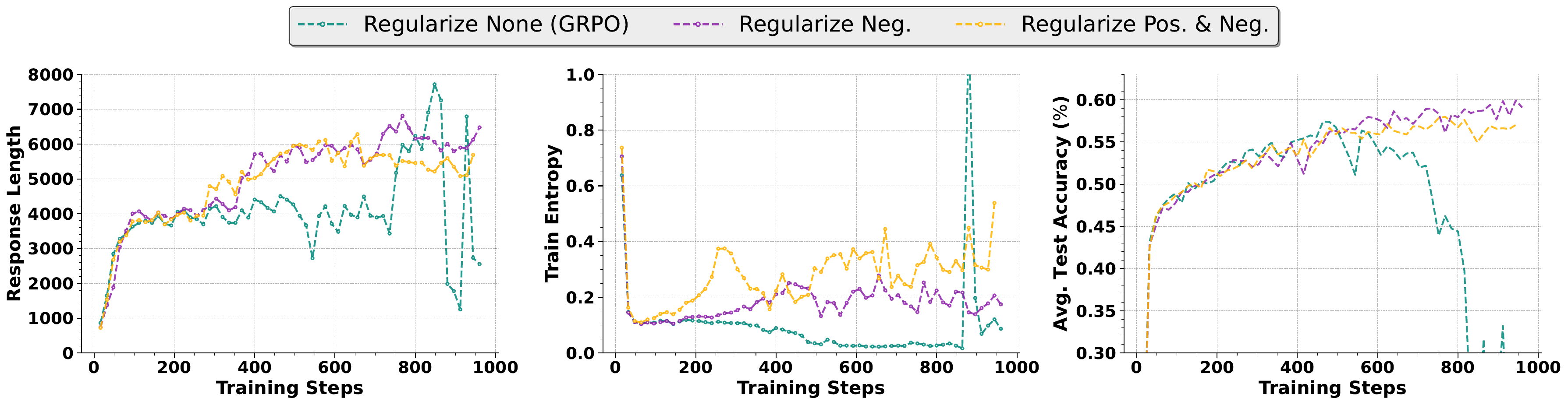}
    \vspace{-20pt}
    \caption{Ablation study comparing \textbf{positive sample regularization and negative sample regularization} for the Lp-Reg penalty (on-policy) on the Qwen3-14B-Base model. Negative sample regularization exhibits better performance.}    \label{fig:ablation_pos_neg_combined_metrics}
\vspace{-5pt}
\end{figure}

\paragraph{Effect of Negative Samples.}
To dissect the contribution of our regularization strategy, we analyze the effects of applying the Lp-Reg penalty to negative samples versus both positive and negative samples. As illustrated in Figure~\ref{fig:ablation_pos_neg_combined_metrics}, applying regularization to low-probability tokens in all samples (Regularize Pos. \& Neg.) sustains exploration and yields continuous performance growth, outperforming standard GRPO (Regularize None). This finding validates that protecting low-probability tokens is essential for successful exploration in RLVR.
Furthermore, the comparison reveals that concentrating the regularization solely on negative samples leads to a faster learning rate than applying it to both sample types. We attribute this to the primary function of Lp-Reg, which is to protect valuable low-probability tokens from being excessively penalized during training. These crucial tokens are far more prevalent in negative samples. Once these tokens are protected, extending the penalty to positive samples provides little additional exploratory advantage. Instead, it appears to introduce minor noise into the gradient updates, which, while not destabilizing, slows the overall training progress. Thus, we conclude that targeting negative samples exclusively is the most efficient and effective application of our method.

\begin{figure}[!t]
    \centering 
    \includegraphics[width=1\textwidth]{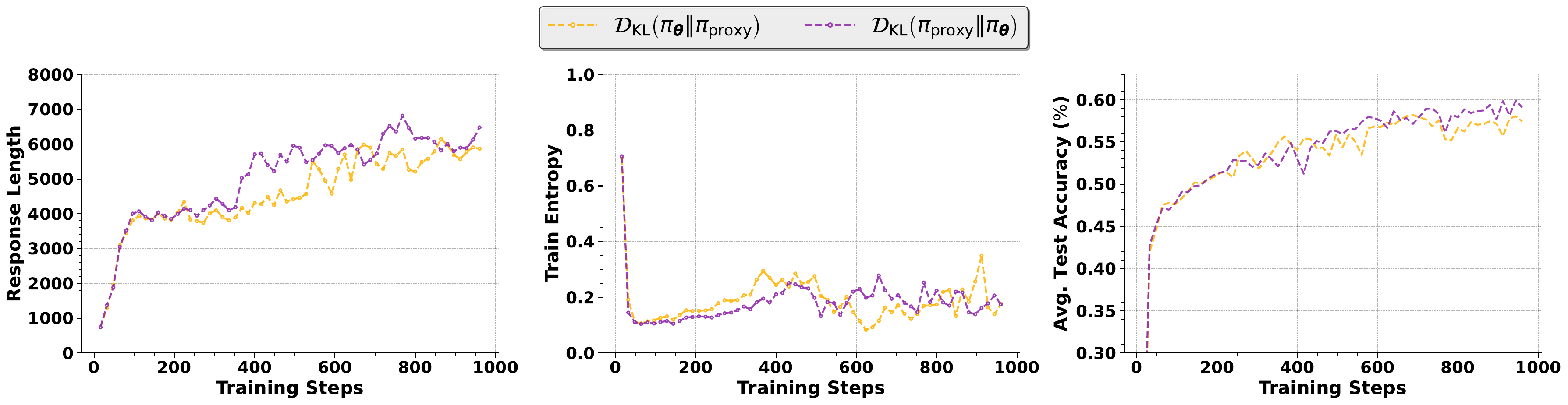}
    \vspace{-20pt}
    \caption{Ablation study comparing the \textbf{forward and reverse KL} formulations for the Lp-Reg penalty (on-policy) on the Qwen3-14B-Base model. The results demonstrate the superiority of the forward KL, which uses the heuristic proxy distribution as a soft guide, over the reverse KL, which forces a strict imitation.}    \label{fig:ablation_forward_reverse_KL_combined_metrics}
\vspace{-5pt}
\end{figure}

\paragraph{Forward KL vs. Reverse KL.}
We compare the performance of our chosen ``forward KL" formulation, which is $\mathcal{D}_\mathrm{KL}(\pi_{\text{proxy}} \Vert \pi_{\boldsymbol{\theta}})$, against the "reverse KL" formulation, $\mathcal{D}_\mathrm{KL}(\pi_{\boldsymbol{\theta}} \Vert \pi_{\text{proxy}})$, in Equation~\ref{eq:lp_reg_objective}. As shown in Figure~\ref{fig:ablation_forward_reverse_KL_combined_metrics}, our forward KL significantly outperforms the reverse KL. This result stems from the fact that our proxy distribution, $\pi_{\text{proxy}}$, is a heuristic reference derived from the current policy, not an ideal target distribution. The reverse KL, $\mathcal{D}_\mathrm{KL}(\pi_{\boldsymbol{\theta}} \Vert \pi_{\text{proxy}})$, penalizes any deviation of $\pi_{\boldsymbol{\theta}}$ from $\pi_{\text{proxy}}$, effectively forcing the policy to strictly imitate this non-ideal, heuristic target. 
This aggressive imitation constrains the protection of potentially valuable exploratory tokens.
In contrast, the forward KL, $\mathcal{D}_\mathrm{KL}(\pi_{\text{proxy}} \Vert \pi_{\boldsymbol{\theta}})$, provides a much softer regularization: it only penalizes the policy for completely discarding tokens that $\pi_{\text{proxy}}$ considers plausible, without forcing a strict match. This allows the policy to use $\pi_{\text{proxy}}$ as a stabilizing guide while retaining the freedom to explore beyond it, which empirically leads to better performance.

% We conduct further ablation studies on the high-entropy token regularization and reverse KL regularization. For detailed results and analysis, please refer to Appendix~\ref{sec:Further Ablation Study}.
\section{Analysis}

To understand the mechanisms behind Lp-Reg's performance, we conduct a series of analyses focusing on how it overcomes the exploration bottleneck by targeting and preserving valuable reasoning tokens.

\subsection{Probability-Entropy Distribution of Exploratory Tokens}\label{sec:Qualitative Analysis of Targeted Tokens}

% \begin{wrapfigure}{r}{0.6\linewidth}
\begin{figure}[!h]
    % \vspace{-0.5cm}
    \centering
    \includegraphics[width=0.8\linewidth]{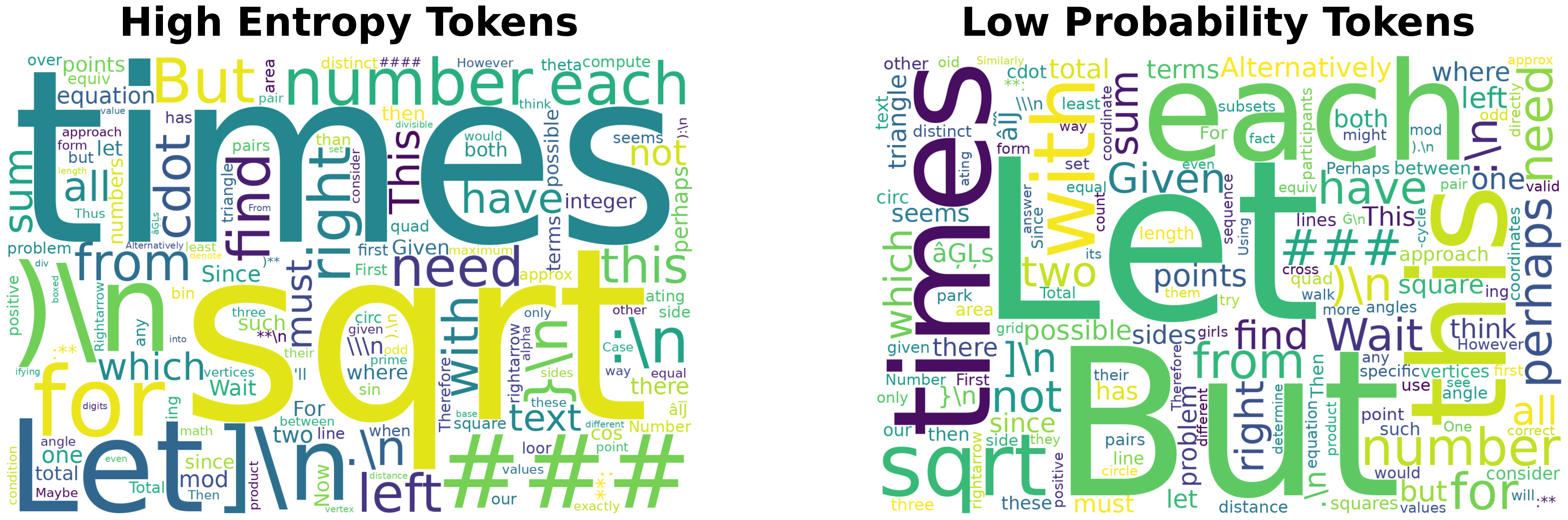}
    \caption{The word cloud statistics from training samples at GRPO training step 1 on
    Qwen3-14B-Base.}
    \label{fig:wordcloud}
    \vspace{-.2cm}
\end{figure}
% \end{wrapfigure}

We begin by exploring the distinction between low-probability tokens and high-entropy tokens. Figure~\ref{fig:wordcloud} highlights this contrast by comparing tokens from the top 1\% lowest probability with those from the top 1\% highest entropy. The difference is striking: low-probability tokens frequently include semantically meaningful exploratory markers such as \emph{``But'', ``Wait'', ``Perhaps'',} and \emph{``Alternatively'',} which often signal a shift in the reasoning trajectory. In contrast, high-entropy tokens are dominated by common functional terms (e.g., \emph{``sqrt'', ``times''}) or formatting symbols (e.g., \texttt{\textbackslash n}), which carry little exploratory intent. This explains why entropy-based regularization often fails to enhance exploration: it confuses noise with exploration.

However, the set of low-probability tokens is also not uniformly useful. It also includes noisy artifacts such as spurious newline characters (\texttt{\textbackslash n}) or formatting debris, whose regularization can destabilize training rather than enhance reasoning. To mitigate this, Lp-Reg applies a threshold $\tau$ that filters out such noise. Ablation studies in Section~\ref{sec:Ablation Study} confirm the necessity of this step: removing the threshold results in unstable training dynamics and degraded reasoning performance. Thus, Lp-Reg’s effectiveness stems not only from targeting low-probability tokens but also from selectively excluding irrelevant noise.

\subsection{Sampling Dynamics of Exploratory Tokens}

\begin{figure}[h!]
\centering
\includegraphics[width=0.9\linewidth]{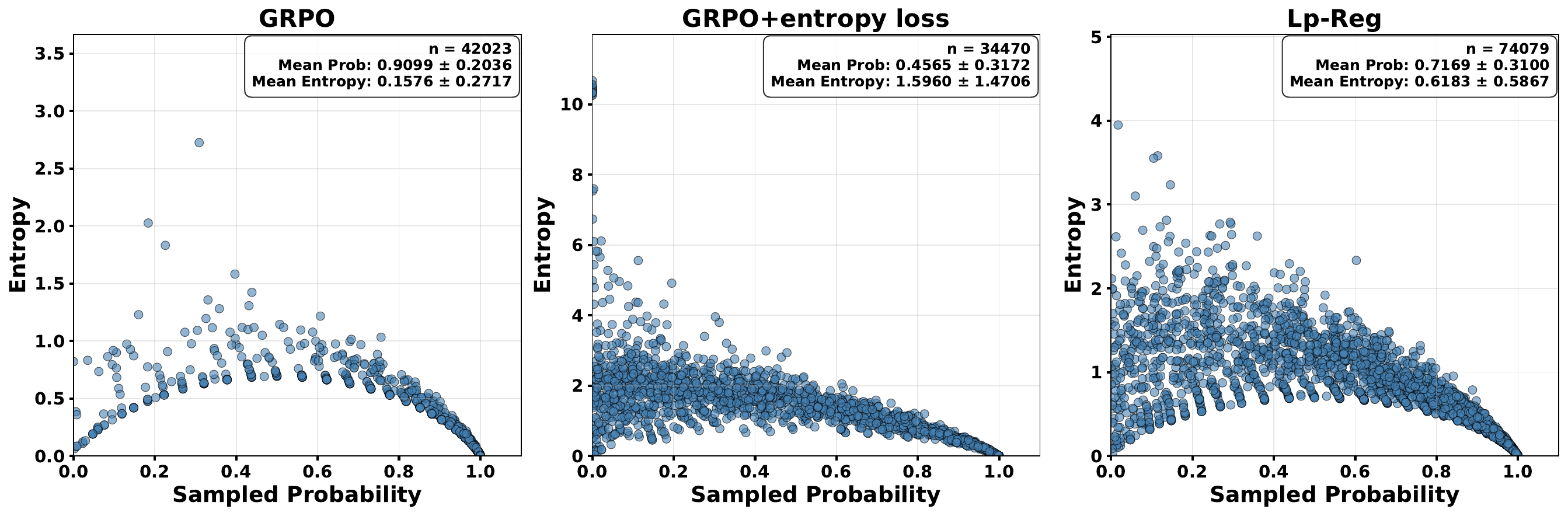}
\caption{Probability–Entropy scatter plots of five explorative tokens from training samples at training step 256 on Qwen3-14B-Base, displaying a random sample of $5\%$ of all data points. Detailed plots for individual tokens are analyzed in Appendix~\ref{sec:Details of Probability-Entropy Distribution}}
\label{fig:scatter}
\end{figure}

Figure~\ref{fig:scatter} shows the probability–entropy distributions of key explorative tokens (“but”, “wait”, “perhaps”, “alternatively”, and “however”) under three methods: GRPO, GRPO + Entropy Loss, and our Lp-Reg.

With the baseline GRPO, these tokens are concentrated in low-entropy, high-probability regions. In this case, tokens like “wait” tend to appear only when the model is already confident, turning them into deterministic patterns rather than initiating a new exploration path with uncertainty.

Adding an entropy loss changes this behavior, but in an uncontrolled way. Some sampled “wait” tokens appear at extremely high entropy levels (sometimes exceeding 10), which superficially boosts diversity but produces little useful exploratory signal. These scattered occurrences do not integrate meaningfully into the reasoning process.

Our Lp-Reg method yields a more balanced dynamic. Explorative tokens are sampled across a broad range of entropy values, from high probability to low probability states. This balance prevents their probabilities from collapsing under negative feedback while keeping them informative for reasoning. As a result, tokens like “wait” remain viable options throughout training, allowing the model to explore alternative reasoning paths rather than overfitting to fixed usage patterns.

\begin{figure}[h!]
\centering
\includegraphics[width=0.95\linewidth]{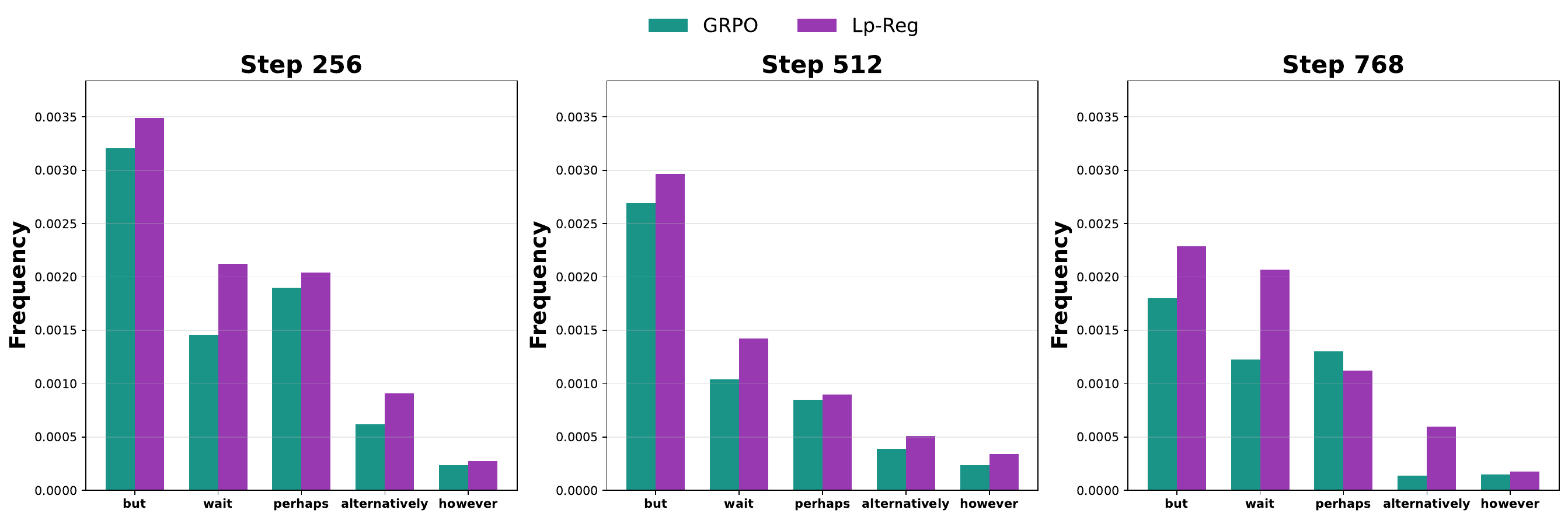}
\caption{Frequency of explorative tokens during training.}
\label{fig:freq}
\end{figure}

Figure~\ref{fig:freq} further compares the frequency of explorative tokens (“but”, “wait”, “perhaps”, “alternatively”, and “however”) under GRPO and Lp-Reg. Our method consistently maintains a higher fraction of these tokens, demonstrating that Lp-Reg not only broadens their probability–entropy distribution but also sustains their practical use throughout training.

\subsection{Probabilistic Distinction between Reasoning Sparks and Noise}

Our introduction established a challenge for a successful exploration strategy: it must protect valuable, low-probability \textit{reasoning sparks} without simultaneously amplifying the destructive effects of irrelevant noise. This raises a critical question: is there a systemic, observable difference between these two classes of tokens within the low-probability range that our method can exploit?

\begin{wrapfigure}{r}{0.5\linewidth}

    \centering
    \includegraphics[width=1.0\linewidth]{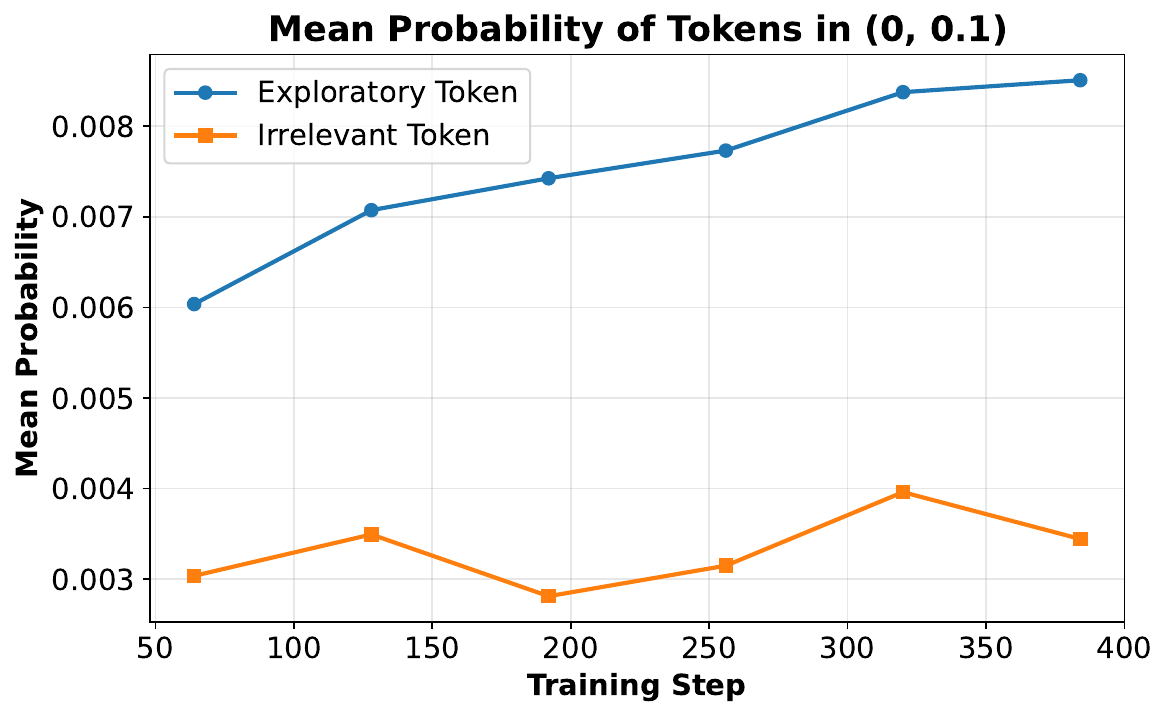}
    \caption{Probabilistic distinction between exploratory and irrelevant tokens across training steps in standard GRPO training.}
    \label{fig:spark_vs_noise}
    \vspace{-.3cm}
\end{wrapfigure}

To investigate this, we analyze the next-token prediction distribution throughout the training process. Due to storage limitations, we focus our analysis on the top-64 most probable tokens, but specifically examine those within a low-probability range ($0$ to $0.1$) to isolate the phenomenon from high-probability tokens. Figure~\ref{fig:spark_vs_noise} plots the average probability of two distinct classes of tokens over time: a group of meaningful exploratory tokens (e.g., ``wait'', ``perhaps'') and a group of irrelevant tokens (e.g., ``cost'', ``fine'').

The results reveal a clear and consistent statistical distinction: across all training stages, the average next-token probability of meaningful exploratory tokens is persistently higher than that of irrelevant tokens. It can be attributed to the intrinsic confidence of LLMs~\citep{nguyen2025turningheatminpsampling, xu2025adaptiveterminationmultiroundparallel, fu2025deepthinkconfidence}. This persistent probabilistic gap provides the foundational justification for our Lp-Reg design. It suggests that while a perfect separation is not possible, a probability threshold $\tau$, as defined for our proxy distribution in Section~\ref{sec:Proxy Distribution}, can serve as a principled filtering mechanism. By setting such a threshold, we can effectively filter out a substantial portion of the lowest-probability irrelevant tokens, which constitute destabilizing noise, while simultaneously retaining a majority of the valuable exploratory tokens that give rise to \textit{reasoning sparks}. This allows Lp-Reg to focus its regularization on tokens that are more likely to be meaningful, providing a targeted and robust approach to preserving high-quality exploration.

\section{Conclusion}

In this work, we investigated the exploration collapse in Reinforcement Learning with Verifiable Rewards. We identified a key mechanism driving this failure: the systematic elimination of a class of valuable, low-probability exploratory tokens we term \textit{reasoning sparks}. To address this, we introduced Low-probability Regularization (Lp-Reg), a method designed to selectively preserve these crucial exploratory pathways. Lp-Reg is founded on the key insight that within the low-probability range, meaningful exploratory tokens consistently exhibit higher average probabilities than semantically irrelevant tokens, whose low-probability appearances act as noise. By leveraging this statistical distinction to filter out irrelevant noise and regularizing the policy towards the remainder, our method effectively protects valuable low-probability tokens from being extinguished. This focus on exploration quality over quantity enables continuous on-policy scaling for around $3,000$ steps, resulting in an absolute $2.66\%$ test accuracy improvement over baselines and underscoring the importance of preserving the \textit{reasoning sparks} within the policy's low-probability tail.

\bibliography{iclr2026_conference}
\bibliographystyle{iclr2026_conference}

\newpage
\appendix
\begin{center}
    \noindent\rule{\textwidth}{4pt} \vspace{-0.2cm}
    \LARGE \textbf{Appendix} % \\ ~\\[-0.5cm]
    \noindent\rule{\textwidth}{1.2pt}
\end{center}
% \wl{why new page here and also for different sections?}

\startcontents[sections]
\printcontents[sections]{l}{1}{\setcounter{tocdepth}{2}}

% \section{Large Language Models Usage Statement}

% In adherence to the ICLR 2026 policy, we disclose the use of a large language model (LLM) as a general-purpose writing assistant during the preparation of this manuscript. The LLM's role was strictly limited to improving the clarity, grammar, and readability of our author-written text, such as spell-checking and rephrasing sentences for better flow. Crucially, the LLM did not contribute to any of the core scientific aspects of this work, including research ideation, experimental design, data analysis, or the generation of novel insights. The authors have carefully reviewed all LLM-modified text and take full responsibility for the intellectual substance and final content of this paper.

\section{Details of Experiments}
% \subsection{Further Training Dynamics} \label{sec:Further Training Dynamics}
% The training dynamics of Lp-Reg and other RLVR methods on the Qwen2.5-32B base model are presented in Figure~\ref{fig:32b-main-results}. The results show that Lp-Reg maintains a comparable performance in test accuracy throughout the training process, underscoring the benefits of our low-probability token regularization strategy for preventing exploration collapse.

% \subsection{Further Ablation Study} \label{sec:Further Ablation Study}

\begin{figure}[h!]
    \centering
    
    \subfloat[Effect of different $\rho$ which defined the low-probability percentile threshold $\delta_{\rho}^{\mathcal{B}}$.]{\label{fig:hyperparam_rho_combined_metrics}
    \includegraphics[width=1\linewidth]{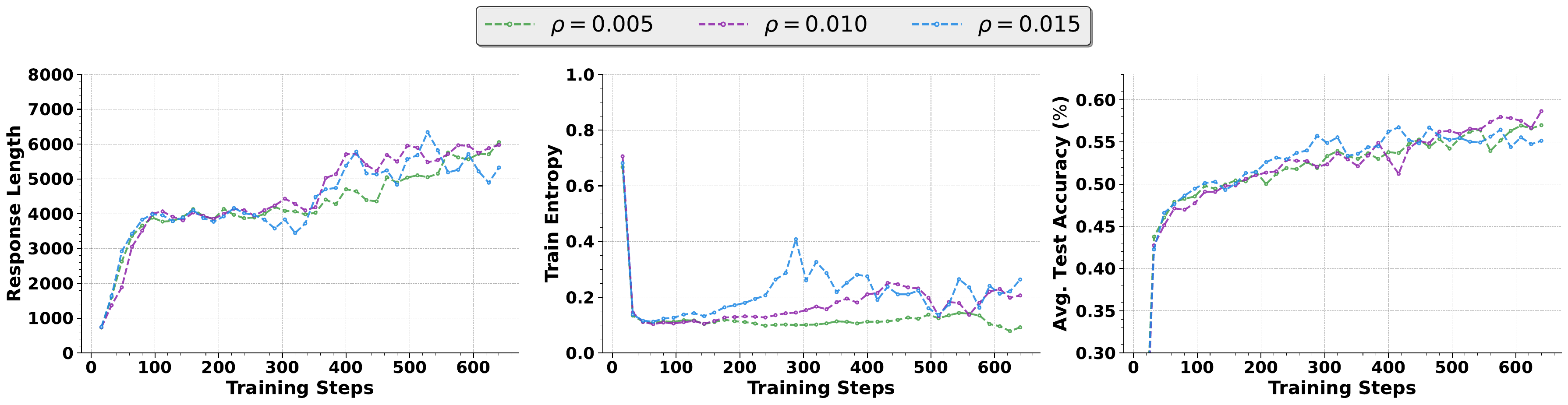}}

    \subfloat[Effect of different $\kappa$ which defined the noise threshold $\tau = \kappa \cdot \max_{o' \in V} \pi_{\boldsymbol{\theta}}(o'|\cdot)$]{\label{fig:hyperparam_kappa_combined_metrics}
    \includegraphics[width=1\linewidth]{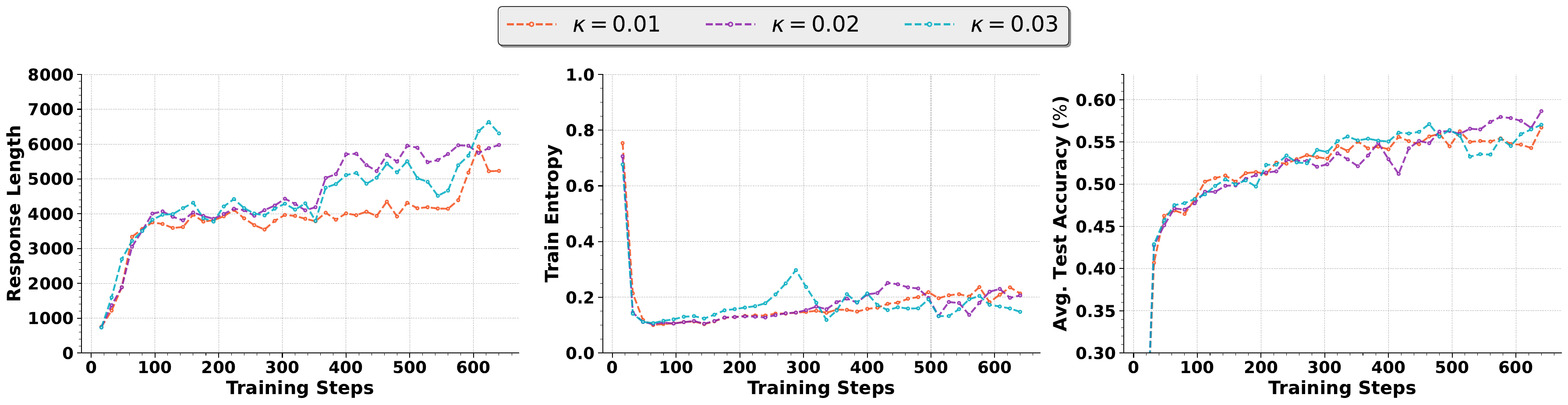}}

    \caption{Training dynamics of Lp-Reg method with different hyperparameters.}
    \label{fig:hyperparameter_sensitivity_analysis}
\end{figure}

\subsection{Hyperparameter Sensitivity Analysis} \label{sec:Hyperparameter Sensitivity Analysis}

In this section, we analyze the sensitivity of two core hyperparameters in Lp-Reg to demonstrate the robustness of our method: the low-probability percentile $\rho$ and the min-p ratio $\kappa$. The results are presented in Figure~\ref{fig:hyperparameter_sensitivity_analysis}.

The parameter $\rho$, as defined in our objective function (Equation~\ref{eq:lp_reg_objective}), determines the percentile threshold for identifying low-probability tokens that are candidates for regularization. A higher $\rho$ means a wider range of tokens are protected. As shown in the top panel of Figure~\ref{fig:hyperparameter_sensitivity_analysis}, we evaluated $\rho$ with values of $0.005$, $0.010$, and $0.015$. The training trajectories for average test accuracy are comparable, and the final performance across all three settings is highly comparable. This indicates that Lp-Reg is not overly sensitive to the precise scope of tokens being protected within this reasonable range.

The hyperparameter $\kappa$ controls the adaptiveness of the min-p filtering threshold, which defines the boundary for what is treated as noise. A smaller $\kappa$ results in a more conservative filtering strategy, removing fewer tokens. Our sensitivity analysis for $\kappa$, presented in the bottom panel of Figure~\ref{fig:hyperparameter_sensitivity_analysis}, shows a similar trend of stability. Across the tested values of $0.01$, $0.02$, and $0.03$, the training curves and final performance remain consistently high and tightly clustered. Taken together, these results demonstrate the robustness of Lp-Reg. The method's effectiveness is not contingent on extensive, fine-grained hyperparameter tuning, highlighting its practical applicability.

\section{Further Analysis}

\subsection{Details of Sampling Probability Density} \label{sec:Details of Sampling Probability Density}

This section provides a detailed, token-by-token breakdown of the aggregated distributions presented in Figure~\ref{fig:c} and Figure~\ref{fig:d} of the main paper, reinforcing the conclusions drawn from our analysis.

\begin{figure}[!h]
    \centering
    
    \subfloat[Density of observed sampling probabilities for token ``but".]{\label{fig:combined_probability_distribution_but}
    \includegraphics[width=1\linewidth]{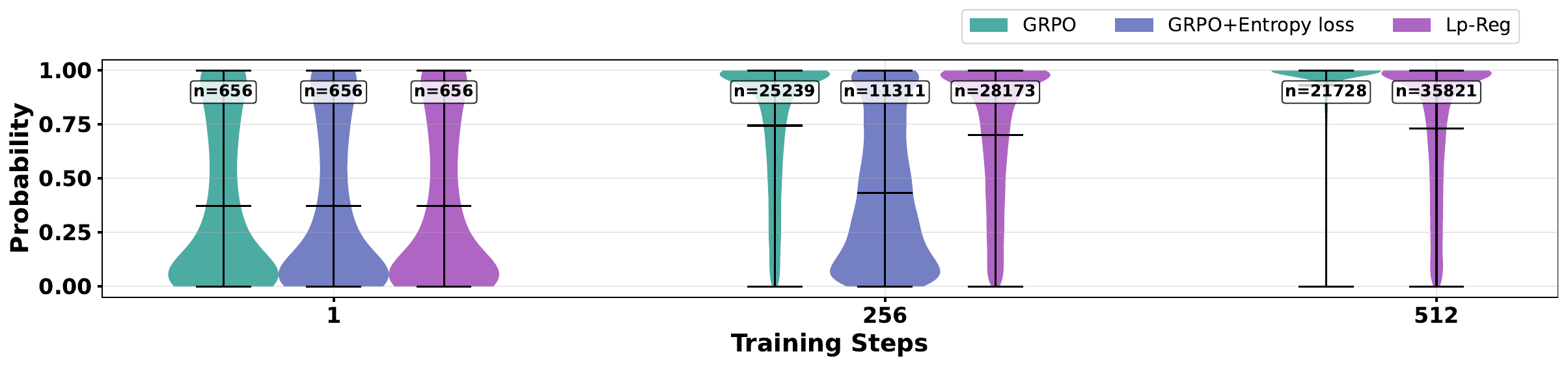}}
    
    \subfloat[Density of observed sampling probabilities for token ``wait".]{\label{fig:combined_probability_distribution_wait}
    \includegraphics[width=1\linewidth]{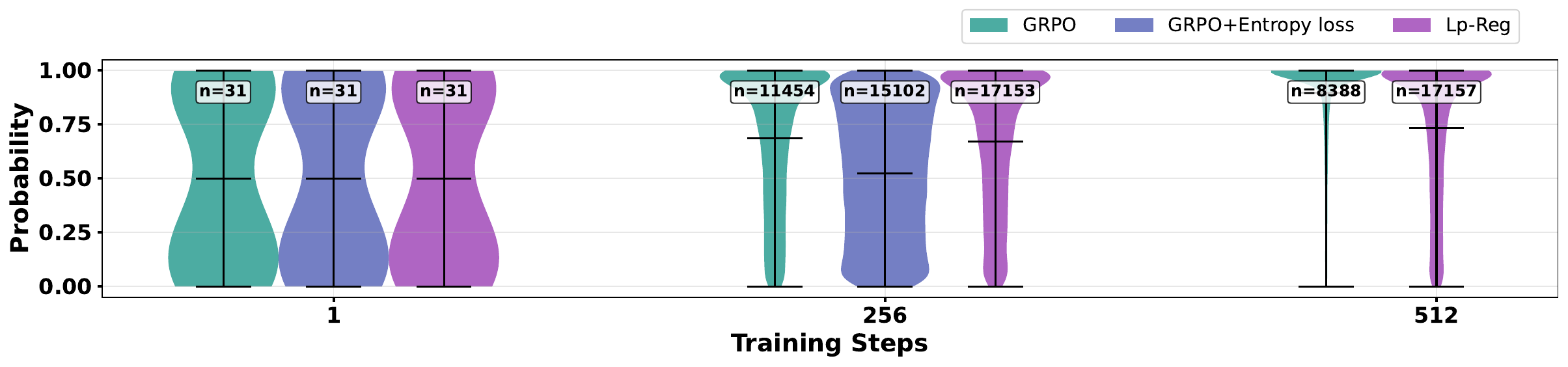}}

    \subfloat[Density of observed sampling probabilities for token ``perhaps".]{\label{fig:combined_probability_distribution_perhaps}
    \includegraphics[width=1\linewidth]{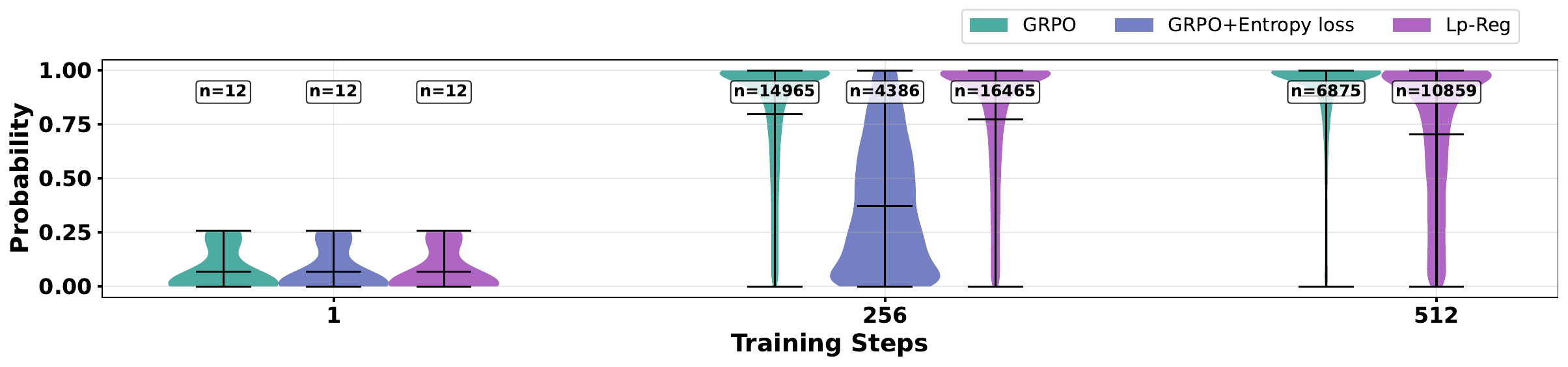}}
    
    \subfloat[Density of observed sampling probabilities for token ``alternatively".]{\label{fig:combined_probability_distribution_alternatively}
    \includegraphics[width=1\linewidth]{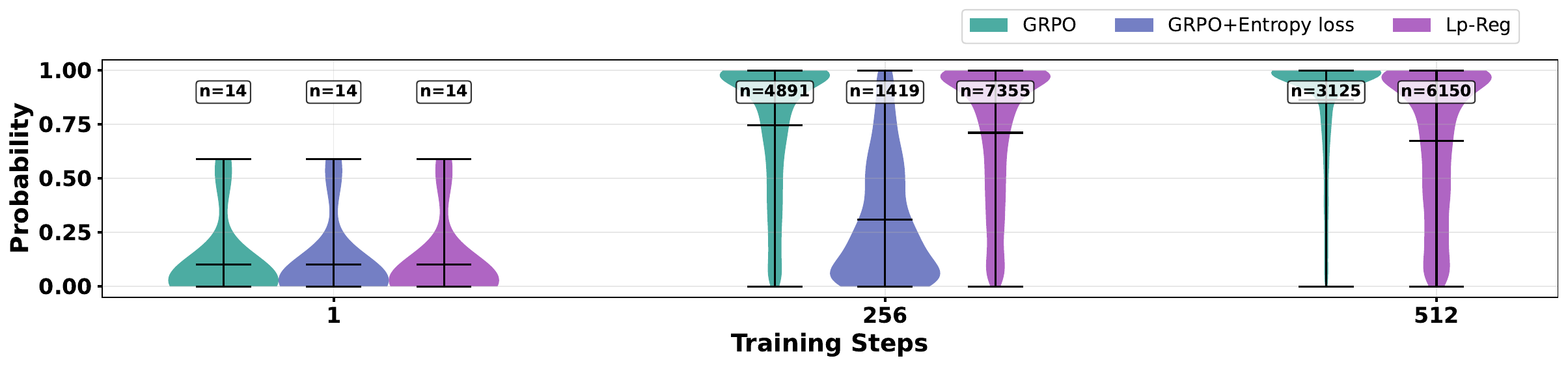}}
    
    \subfloat[Density of observed sampling probabilities for token ``however".]{\label{fig:combined_probability_distribution_however}
    \includegraphics[width=1\linewidth]{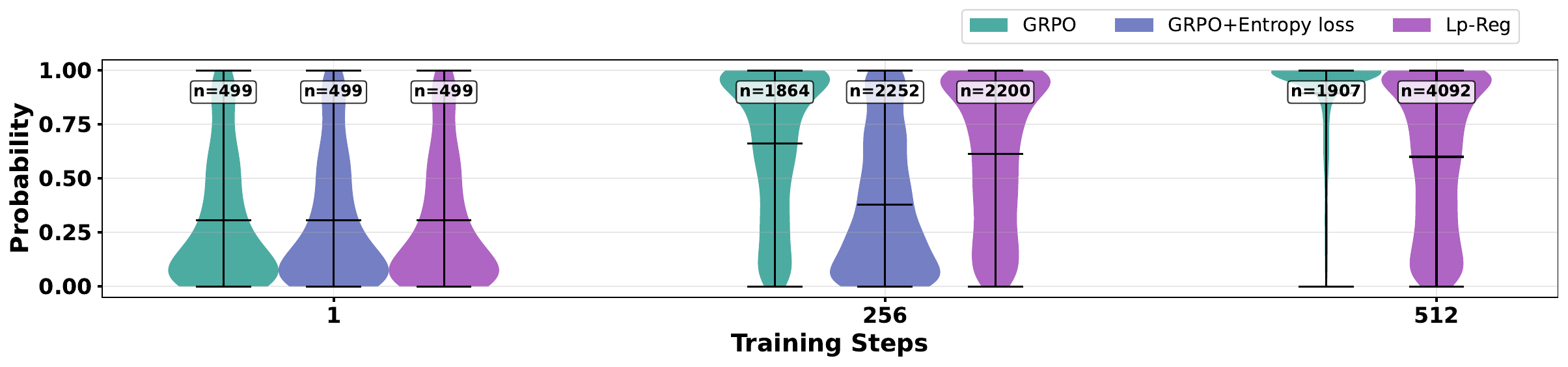}}
    
    \caption{Individual density of observed sampling probabilities for meaningful exploratory tokens: ``but", ``wait", ``perhaps", ``alternatively", and ``however".}
    \label{fig:meaningful_all_combined_distributions} 
\end{figure}

Figure~\ref{fig:meaningful_all_combined_distributions} exhibits the individual distribution of observed sampling probabilities for a class of meaningful low-probability exploratory tokens we term \textit{reasoning sparks}: ``but'', ``wait'', ``perhaps'', ``alternatively'', and ``however''. A consistent trend is observable across all five tokens, validating our claims in the introduction. With standard GRPO training, the ability to sample these tokens at low probabilities is systematically eliminated, causing their distributions to collapse and shift towards higher probabilities. The indiscriminate entropy bonus (GRPO + Entropy Loss) is largely ineffective at restoring this crucial low-probability tail. In stark contrast, our proposed method, Lp-Reg, consistently maintains a healthy, wide distribution for each of these tokens, demonstrating its effectiveness in preserving the model's capacity for exploration.

\begin{figure}[h!]
    \centering

    \subfloat[Density of observed sampling probabilities for token ``cost".]{\label{fig:combined_probability_distribution_cost}
    \includegraphics[width=1\linewidth]{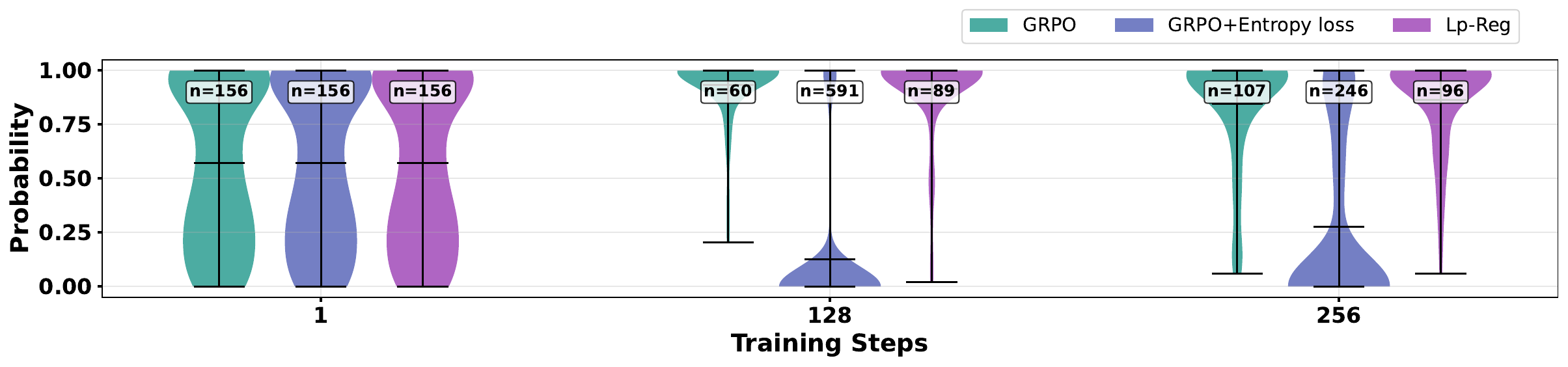}}

    \subfloat[Density of observed sampling probabilities for token ``fine".]{\label{fig:combined_probability_distribution_fine}
    \includegraphics[width=1\linewidth]{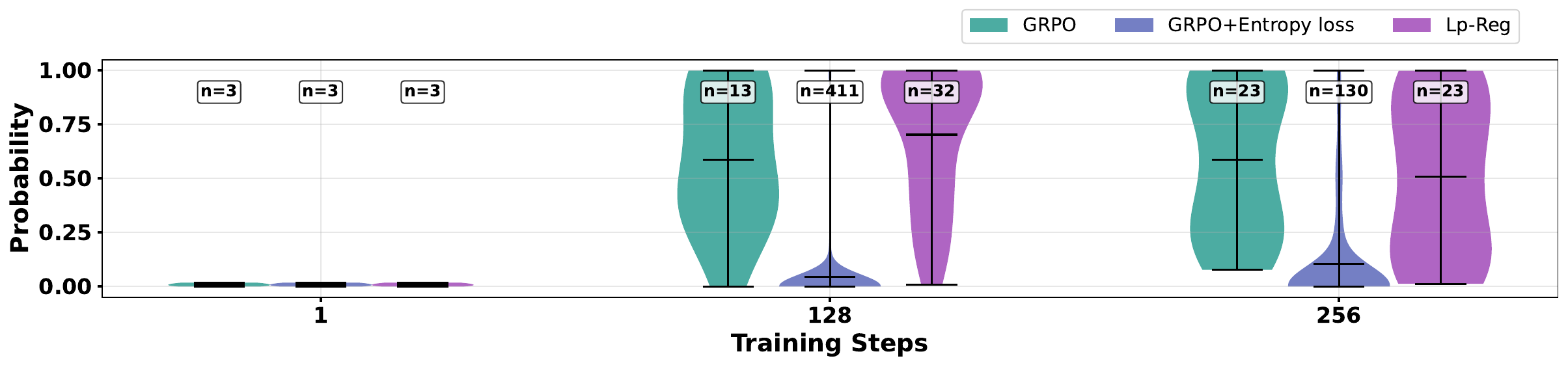}}

    \subfloat[Density of observed sampling probabilities for token ``balanced".]{\label{fig:combined_probability_distribution_balanced}
    \includegraphics[width=1\linewidth]{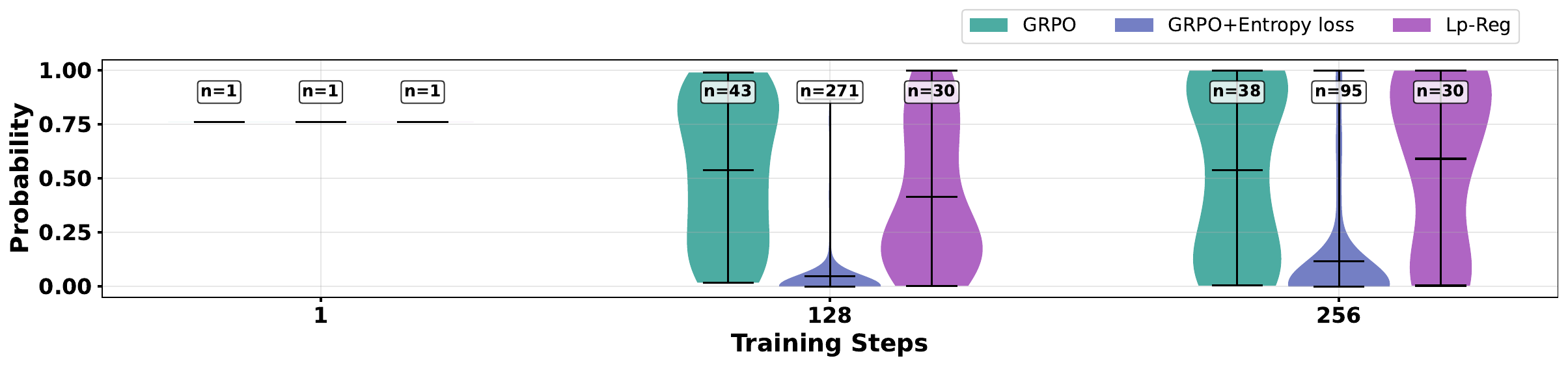}}

    \subfloat[Density of observed sampling probabilities for token ``ere".]{\label{fig:combined_probability_distribution_ere}
    \includegraphics[width=1\linewidth]{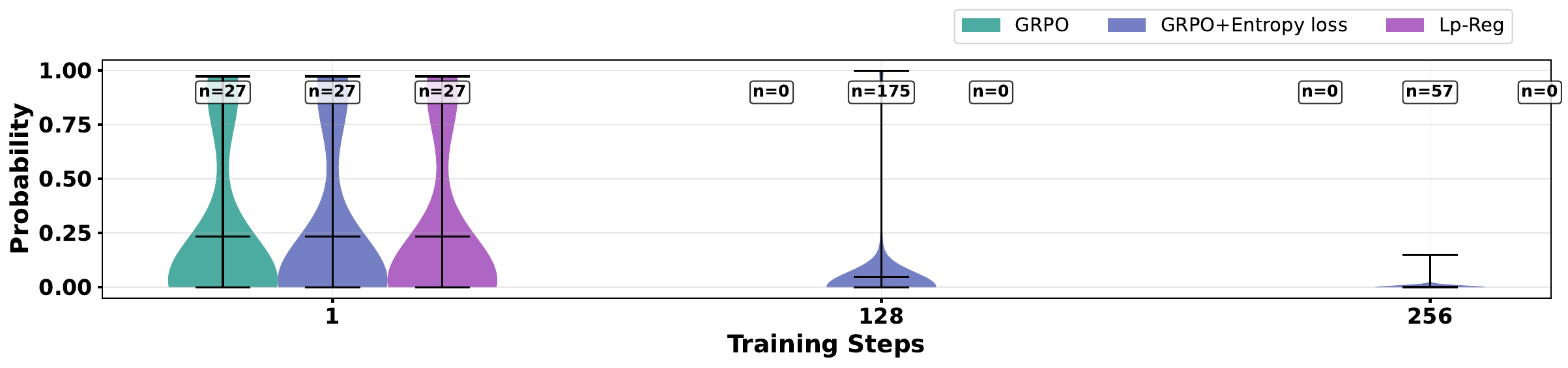}}

    \subfloat[Density of observed sampling probabilities for token ``trans".]{\label{fig:combined_probability_distribution_trans}
    \includegraphics[width=1\linewidth]{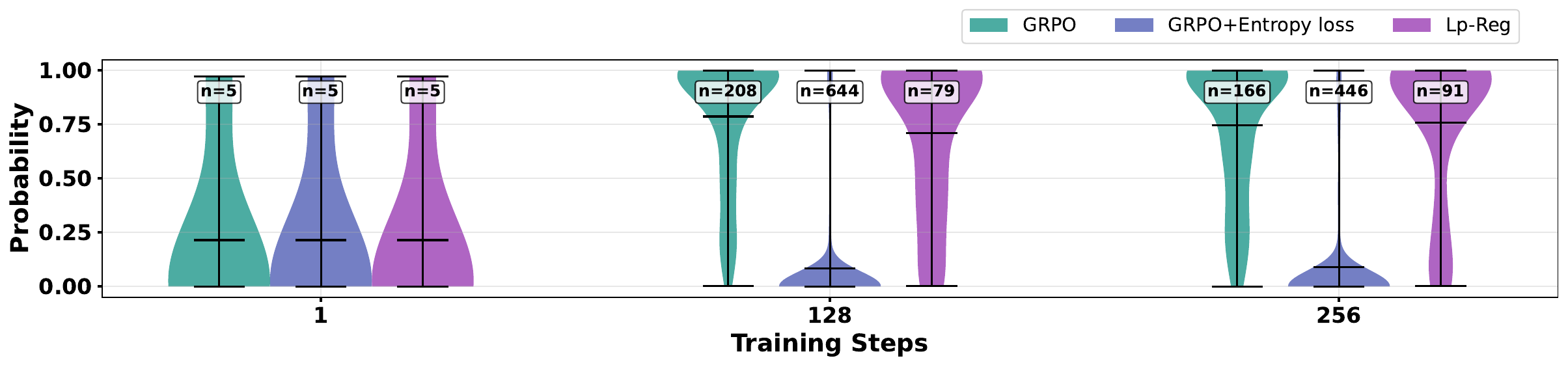}}

    \caption{Individual density of observed sampling probabilities for irrelevant tokens: ``cost", ``fine", ``balanced", ``ere", and ``trans".}
    \label{fig:meaningless_all_combined_distributions} 
\end{figure}

Conversely, Figure~\ref{fig:meaningless_all_combined_distributions} details the behavior of a class of what we term irrelevant tokens (e.g. ``cost'', ``fine'', ``balanced'', ``ere'', and ``trans''). We refer to the low-probability sampling of these tokens as irrelevant noise, which can be destructive to the training process. These individual plots clearly illustrate the detrimental side effect of a simple entropy bonus. For nearly every token, the GRPO + Entropy Loss baseline significantly amplifies the sampling of this irrelevant noise, which, as shown in our main analysis, contributes to a faster performance collapse. Lp-Reg, by design, avoids this amplification and maintains a suppressed probability distribution for these tokens, comparable to or even more constrained than the standard GRPO baseline.

These detailed visualizations confirm that the elimination of \textit{reasoning sparks} and the amplification of irrelevant noise are not artifacts of aggregation but are consistent patterns at the individual token level. This provides strong, granular evidence for the central challenge our paper addresses and highlights the necessity of a selective preservation mechanism like Lp-Reg.

\subsection{Details of Probability-Entropy Distribution}
\label{sec:Details of Probability-Entropy Distribution}

\begin{figure}[h!]
    \centering

    \subfloat[Scattered probability–entropy plot of observed sampling instances for the token ``but".]{\label{fig:scatter_but}
    \includegraphics[width=0.75\linewidth]{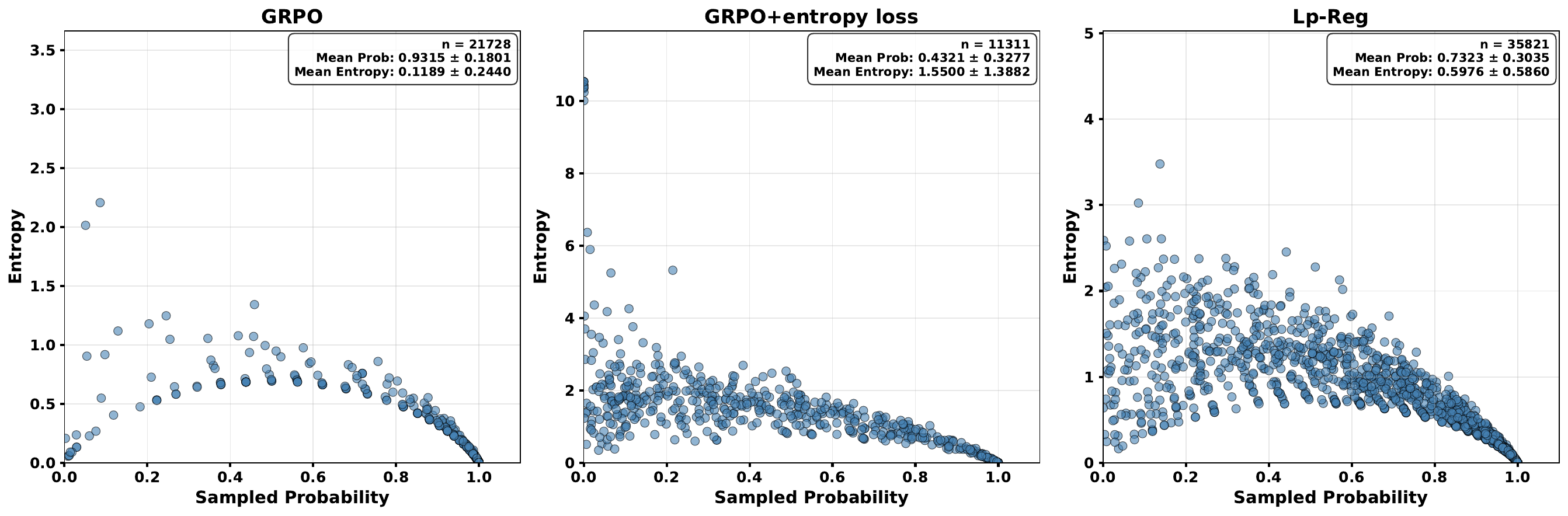}}

    \subfloat[Scattered probability–entropy plot of observed sampling instances for the token ``wait".]{\label{fig:scatter_wait}
    \includegraphics[width=0.75\linewidth]{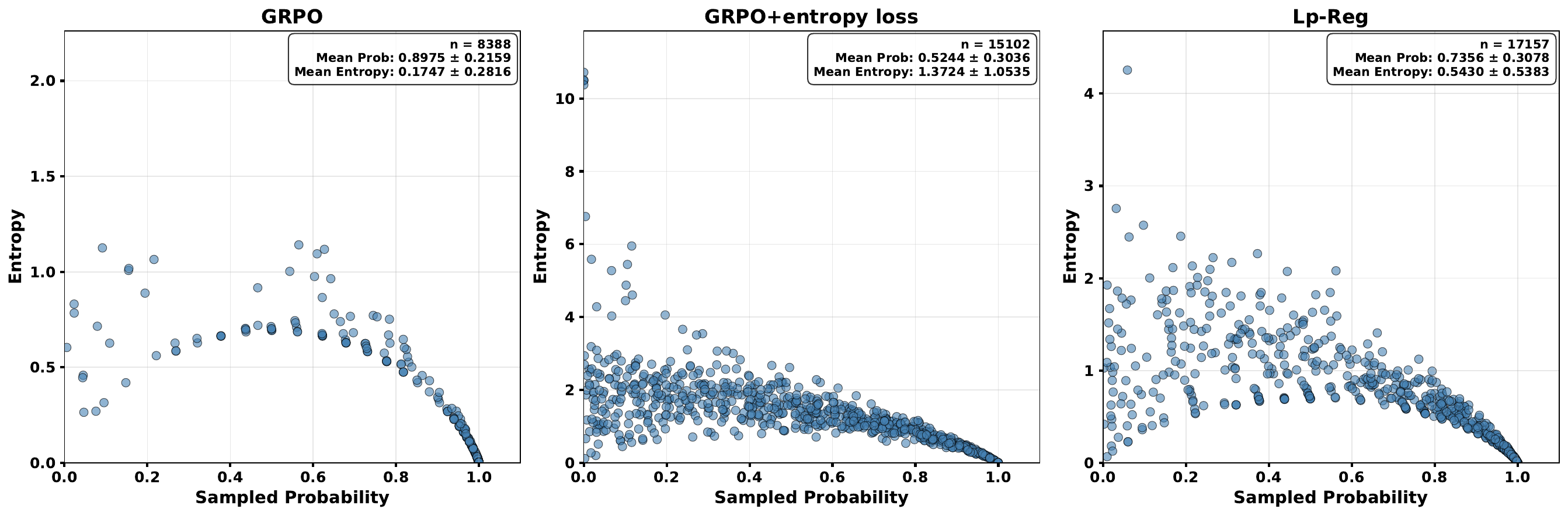}}

    \subfloat[Scattered probability–entropy plot of observed sampling instances for the token ``perhaps".]{\label{fig:scatter_perhaps}
    \includegraphics[width=0.75\linewidth]{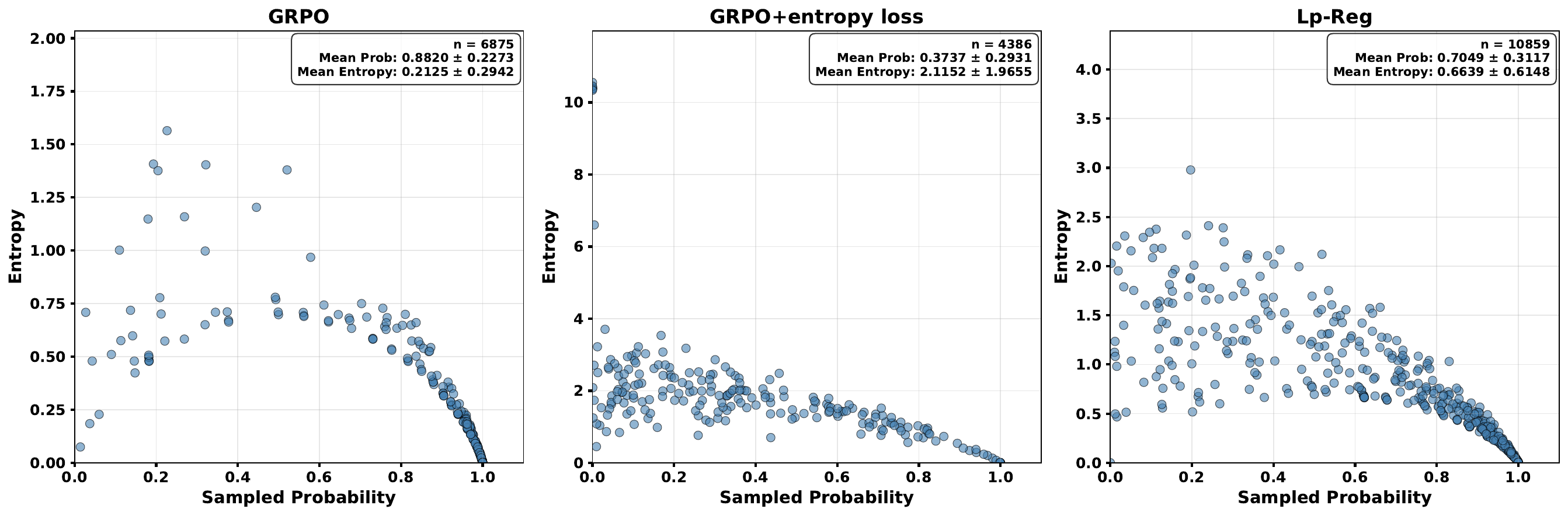}}

    \subfloat[Scattered probability–entropy plot of observed sampling instances for the token ``alternatively".]{\label{fig:scatter_alternatively}
    \includegraphics[width=0.75\linewidth]{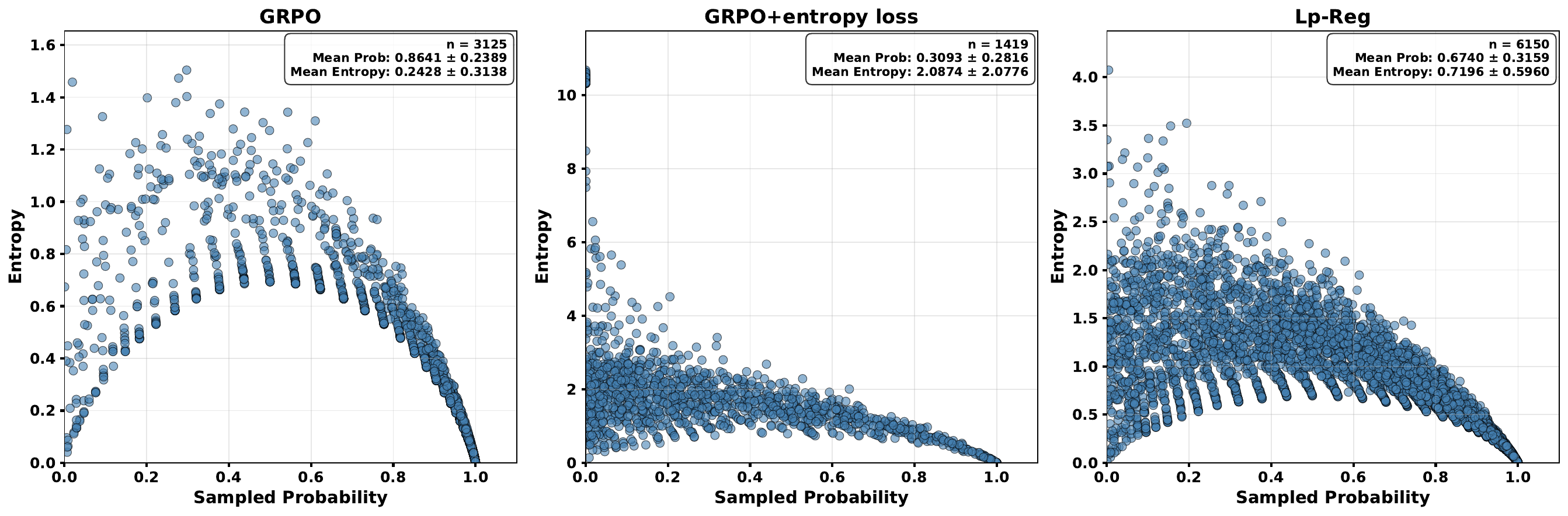}}

    \subfloat[Scattered probability–entropy plot of observed sampling instances for the token ``however".]{\label{fig:scatter_however}
    \includegraphics[width=0.75\linewidth]{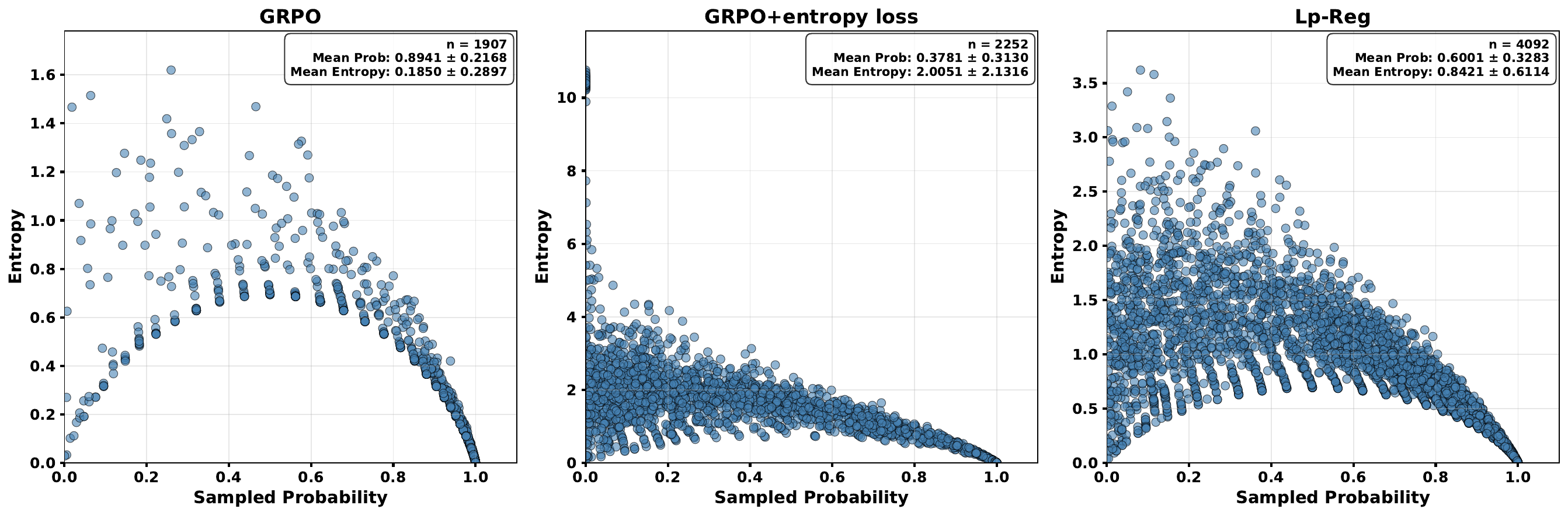}}

    \caption{Individual scattered probability–entropy plot of observed sampling instances for meaningful tokens: ``but", ``wait", ``perhaps", ``alternatively", and ``however".}
    \label{fig:meaningfull_all_scatter} 
\end{figure}

To supplement the aggregated analysis presented in Figure~\ref{fig:scatter} of the main text, this section provides a detailed breakdown of the probability-entropy distributions for individual meaningful exploratory tokens. Figure~\ref{fig:meaningfull_all_scatter} shows a consistent pattern across all representative tokens, ranging from “but” (Figure~\ref{fig:scatter_but}) to “however” (Figure~\ref{fig:scatter_however}). For frequently occurring tokens such as “but”, “wait”, and “perhaps”, we randomly subsample one out of every 20 instances for visualization. Under the baseline GRPO, these sparks are consistently confined to a low-entropy, high-probability region, indicating a collapse into deterministic usage. In contrast, the addition of an entropy loss pushes these tokens into highly scattered, often excessively high-entropy states, suggesting an uncontrolled and potentially noisy form of exploration. Our method, Lp-Reg, strikes a crucial balance, maintaining a structured and broad distribution across a healthy range of entropy values.
This consistent behavior demonstrates that the trends identified in the aggregated data are not artifacts of averaging. The individual plots offer strong, disaggregated evidence for our central claim: Lp-Reg effectively preserves the exploratory potential of reasoning sparks by preventing both the deterministic collapse seen in the baseline and the chaotic scattering induced by the indiscriminate entropy bonus.

\subsection{Training Dynamics of Regularized Token} \label{sec:Training Dynamics of Regularized Token}

\begin{figure}[!h]
    \centering 
    \includegraphics[width=1\textwidth]{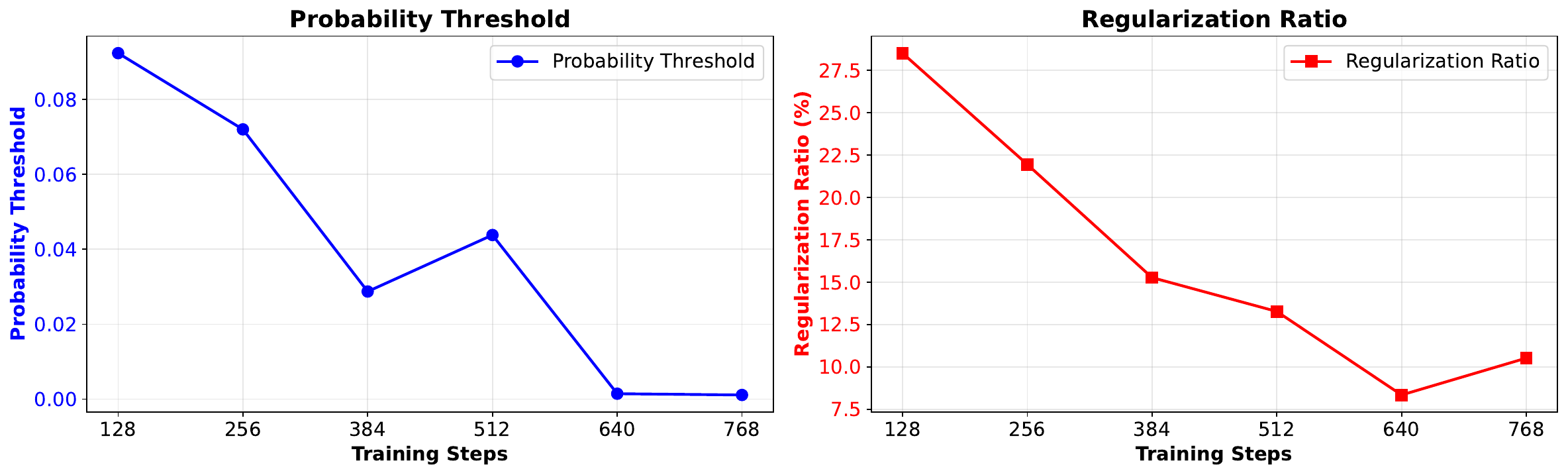}
    \caption{Training dynamics of the probability threshold and regularization ratio.}
    \label{fig:threshold}
\end{figure}

To better understand how Lp-Reg operates during training, we analyze the dynamics of the probability threshold $\delta_{\rho}^{\mathcal{B}}$ and the proportion of low-probability tokens subjected to regularization $\frac{|\pi_{\boldsymbol{\theta}}(o \vert \cdot)<\delta_{\rho}^{\mathcal{B}} \land \pi_{\text{proxy}}(o \vert \cdot) > 0|}{|\pi_{\boldsymbol{\theta}}(o \vert \cdot)<\delta_{\rho}^{\mathcal{B}}|}$. As shown in Figure~\ref{fig:threshold}, the threshold $\delta_{\rho}^{\mathcal{B}}$ gradually decreases with training steps. At the same time, the regularization ratio also declines steadily. This trend suggests that as training progresses, the extreme low-probability range becomes increasingly dominated by irrelevant tokens, constituting what we term irrelevant noise. It also explains the relatively worse performance using a fixed threshold $\tau=0.02$ than the dynamic min-p threshold $\tau$ in Equation~\ref{eq:proxy_dist}, since a fixed threshold failed to regularize the lowest $1\%$ tokens while the dynamic one succeeded. Furthermore, the semantically meaningful exploratory tokens are lifted into higher-probability regions, thus requiring less regularization.

\subsection{Case Study}

\begin{figure}
    \centering
    \includegraphics[width=0.9\linewidth]{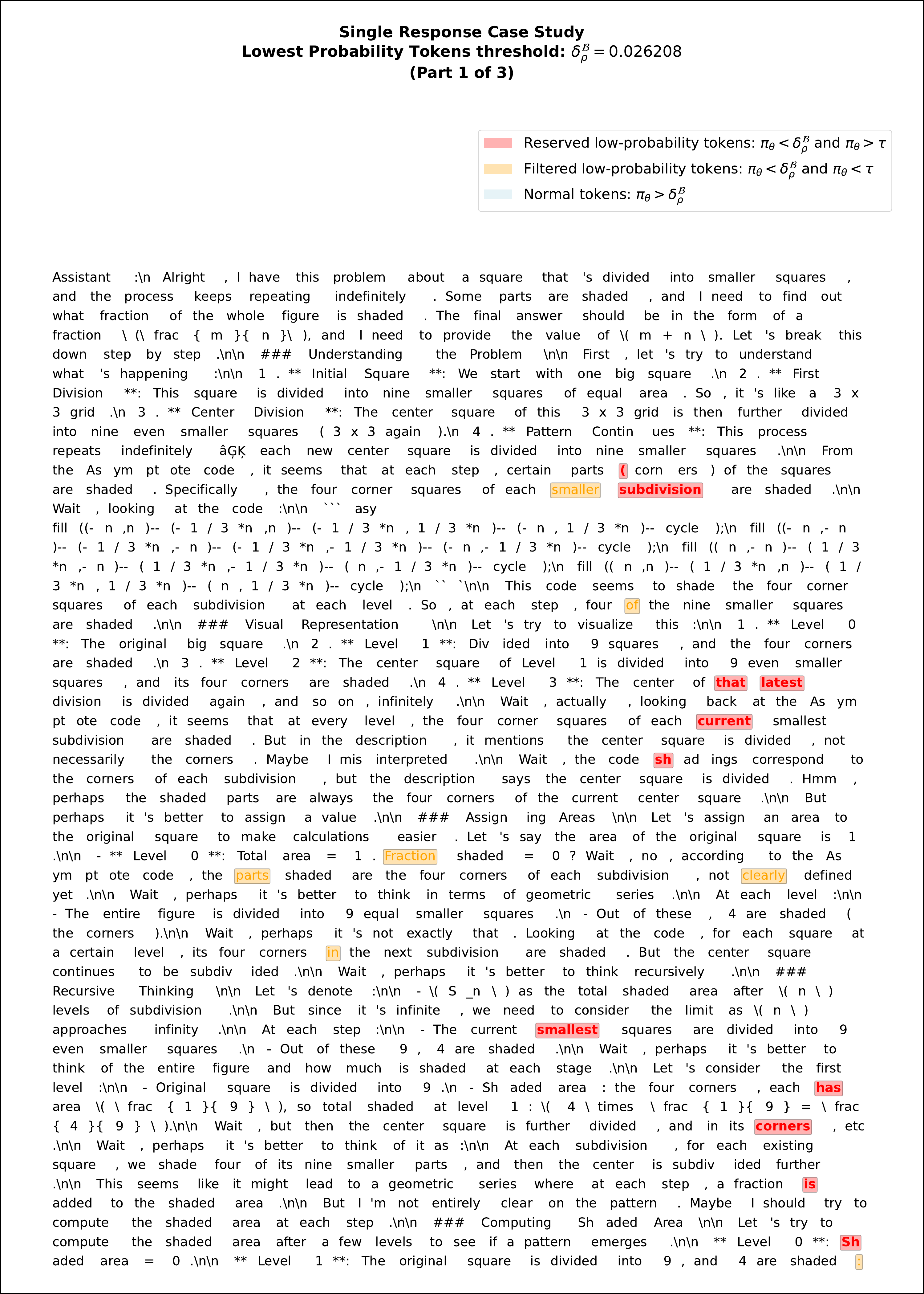}
    \caption{An Example generated by Qwen3-14B-Base model trained by Lp-Reg from math reasoning. (Part 1)}
    \label{fig:case_study_part_1}
\end{figure}

\begin{figure}
    \centering
    \includegraphics[width=0.9\linewidth]{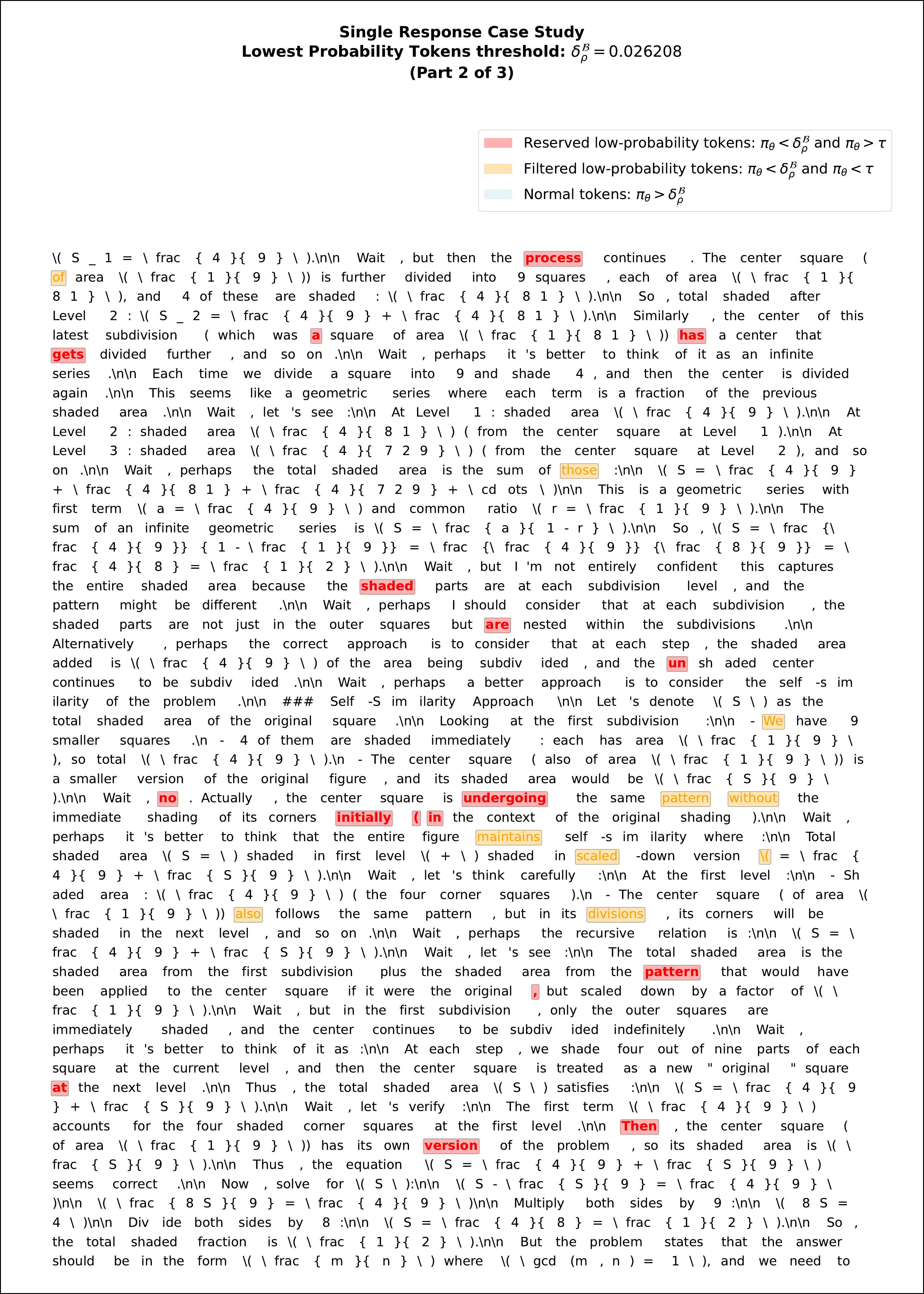}
    \caption{An Example generated by Qwen3-14B-Base model trained by Lp-Reg from math reasoning. (Part 2)}
    \label{fig:case_study_part_2}
\end{figure}

\begin{figure}
    \centering
    \includegraphics[width=0.9\linewidth]{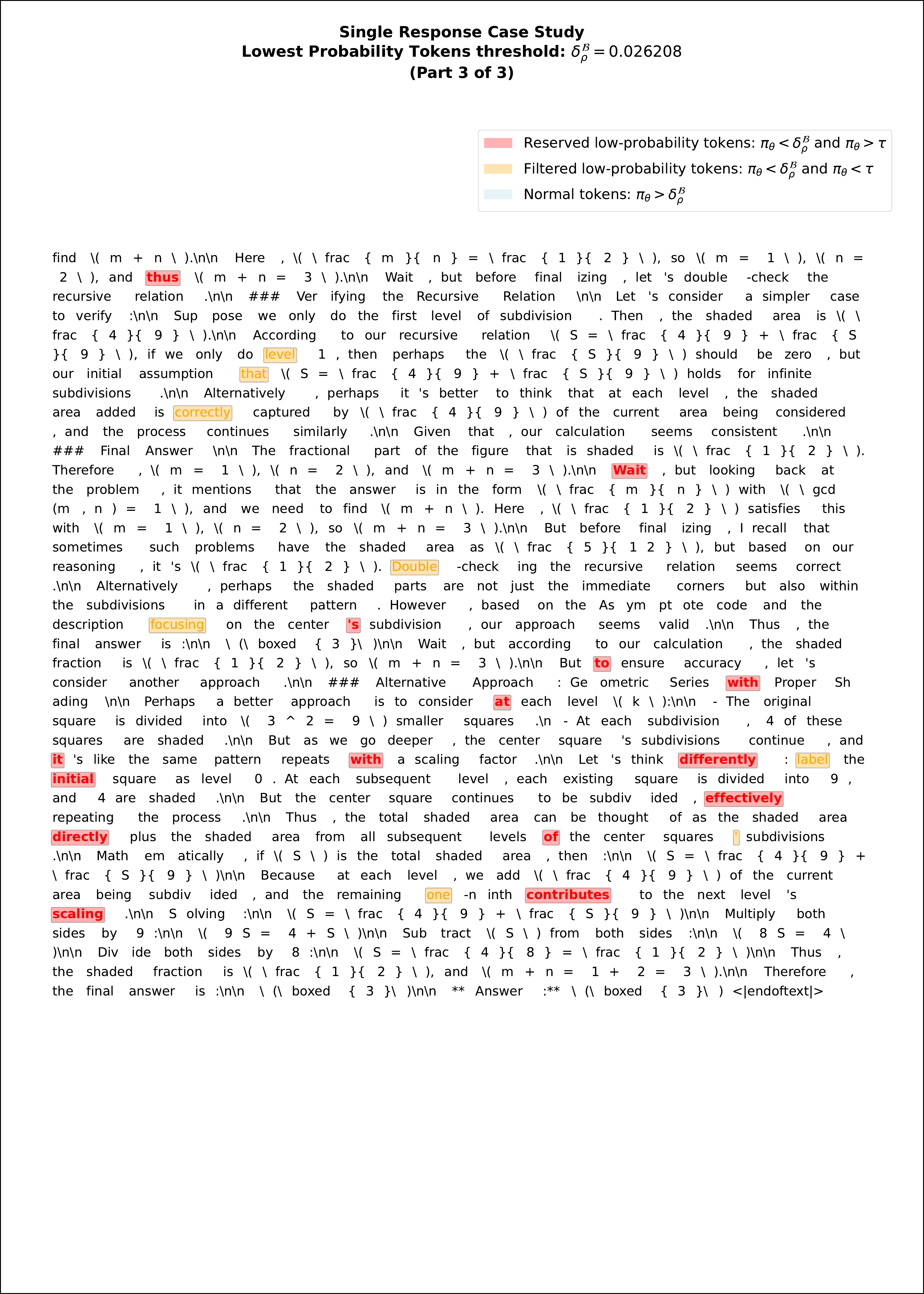}
    \caption{An Example generated by Qwen3-14B-Base model trained by Lp-Reg from math reasoning. (Part 3)}
    \label{fig:case_study_part_3}
\end{figure}

To further illustrate the effect of the filter applied on low-probability tokens, Figure~\ref{fig:case_study_part_1} to Figure~\ref{fig:case_study_part_3} present a case study of a model-generated response, where low-probability tokens are highlighted according to whether they were preserved or filtered. Tokens with probability greater than $\tau$ are those retained by the filter, while tokens with probability smaller than $\tau$ are suppressed. The preserved tokens include meaningful exploratory markers such as "Then", "Wait", which guide the reasoning trajectory, whereas the discarded set largely consists of relatively irrelevant tokens such as "We", "also", "that". This qualitative evidence complements our quantitative analysis, demonstrating that Lp-Reg effectively leverages the min-p transformation to distinguish between valuable low-probability exploratory tokens (\textit{reasoning sparks}) and tokens that constitute destabilizing, irrelevant noise.

\end{document}